\documentclass[journal]{IEEEtran}

\usepackage[showwarnings]{silence}
\usepackage[T1]{fontenc} 
\usepackage[utf8]{inputenc} 
\usepackage[english]{babel} 
\usepackage{amsthm}
\usepackage{amsfonts} 
\usepackage{dsfont} 
\usepackage{gensymb} 
\usepackage{array,multirow,makecell}
\usepackage{cite} 

\usepackage{amsmath}
\usepackage[pdftex]{graphicx}
\graphicspath{
  {pictures/}
  {pictures/teaser/}
  {pictures/patternsWithRectangles/}
  {comp/pictures/}
  {comp/matlab_measures/}
  {comp/matlab_measures/out_EPSfiles/}
  {comp/matlab_measures/out_PNGfiles/}
  {comp/matlab_measures/out_PNGfilesTP/}
  {comp/matlab_testimages/}
  {comp/matlab_testpatterns/}
  {comp/inskape_testpatterns/}
  {comp/lef_matlabfigs/}
  {comp/mgf_matlabfigs/}
  {comp/mbfr_matlabfigs/}
  {comp/bgrain_pictures/}
  {comp/eef_pictures/}}
\DeclareGraphicsExtensions{.pdf,.eps,.png,.jpg}

\usepackage{float}

\usepackage[usenames,dvipsnames,table,hyperref]{xcolor}

\usepackage[ruled,linesnumbered,vlined]{algorithm2e}
\SetKwInOut{Input}{input}
\SetKwInOut{Output}{output}
\SetKwComment{Comment}{}{} 

\SetCommentSty{mycmtsty}
\SetKwProg{Fn}{Function}{}{}

\usepackage[colorlinks,allcolors={APink},pagebackref]{hyperref}

\definecolor{ARed}{RGB}{255,59,48}
\definecolor{AOrange}{RGB}{255,149,0}
\definecolor{AYellow}{RGB}{255,204,0}
\definecolor{AGreen}{RGB}{76,217,100}
\definecolor{ATealBlue}{RGB}{90,200,250}
\definecolor{ABlue}{RGB}{0,122,255}
\definecolor{APurple}{RGB}{88,86,214}
\definecolor{APink}{RGB}{255,45,85}

\newcommand{\dobib}{ 
  \bibliographystyle{IEEEtran}
  \bibliography{IEEEabrv,bibliography/biblio.bib}
}

\newcommand{\dolistoftodos}{
  \listoftodos[List of ToDos]}

\newcommand{\textfg}[1]{\footnotesize{#1}}

\newcommand{\dxo}{DxO} 

\newcommand{\ie}{\textit{i.e.}} 
\newcommand{\etal}{\textit{et al.}} 

\def\x{\textnormal{\bf{x}}}
\def\y{\textnormal{\bf{y}}}

\def\im{u} 
\def\imm{v} 

\def\lpyr{\textnormal{\em Lpyr}} 
\newcommand{\psnr}{\text{PSNR}} 

\def\ffgf{\text{GF}} 
\def\ffbfr{\text{BFR}} 
\def\ffbf{\text{BF}} 
\def\fffbf{\text{FBF}} 
\def\ffdt{\text{DT}} 
\def\ffllf{\text{LLF}} 
\def\fffllf{\text{FLL}} 
\def\ffwls{\text{WLS}} 
\def\fftv{\text{TV}} 
\def\fftv{\text{TV-$L^1$}} 
\def\fflois{\text{IS-$L^0$}} 

\begin{document}

\renewcommand{\dobib}{} 
\renewcommand{\dolistoftodos}{} 

\title{Quantitative Evaluation of Base and Detail Decomposition Filters Based on
  their Artifacts}

\author{
  Charles~Hessel
  and Jean-Michel~Morel%
  \thanks{Charles Hessel and Jean-Michel Morel are with CMLA, ENS Cachan, CNRS,
    Universit\'{e} Paris-Saclay, 94235 Cachan, France}%
}

\markboth{}{}

\maketitle


\begin{abstract}
  This paper introduces a quantitative evaluation of filters that seek to separate an image into its
  large-scale variations, the base layer, and its fine-scale variations, the
  detail layer. Such methods have proliferated  with the development of HDR imaging and the proposition of many new tone-mapping  operators.  We argue that an objective quality measurement for all methods  can be based on their artifacts.
  To this aim, the four  main recurrent artifacts are described and
  mathematically characterized. Among them two are classic, the luminance halo
  and the staircase effect, but  we show the relevance of two more, the contrast
  halo and the compartmentalization effect.
  For
  each of these artifacts we design a  test-pattern  and its attached measurement formula.  Then we fuse these measurements into a single quality mark, and obtain in that  way a ranking method valid for  all filters performing a base+detail decomposition.
  This synthetic ranking is applied to seven filters
  representative of the literature and shown to agree with expert artifact rejection criteria.
\end{abstract}


{\textit{Supplementary Material---}Supplementary material for this paper can
  be found \href{http://dev.ipol.im/~hessel/publications/quantitativeArtifactEvaluation/SupplementaryMaterial.pdf}{here}\footnote{\url{http://dev.ipol.im/~hessel/publications/quantitativeArtifactEvaluation/SupplementaryMaterial.pdf}}.
}

\section{Introduction}
\label{sec:intro}


This paper studies image-processing filters that seek to separate an image
into its large-scale variations, the base layer, and its fine-scale
variations, the detail layer.
Such methods are also referred to as edge-aware smoothing filters or
cartoon+texture decomposition. A base+detail decomposition is the core
of photo editing tools such as high-dynamic range tone mapping and local
contrast enhancement. These algorithms aim at adding more clarity to the image
by enhancing its detail.

\begin{figure*} 
  \centering
  \setlength{\tabcolsep}{.008\linewidth}
  \begin{tabular}{@{}cccc@{}}
    \includegraphics[width=.238\linewidth]{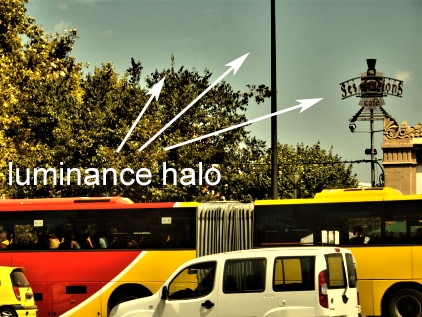} &
    \includegraphics[width=.238\linewidth]{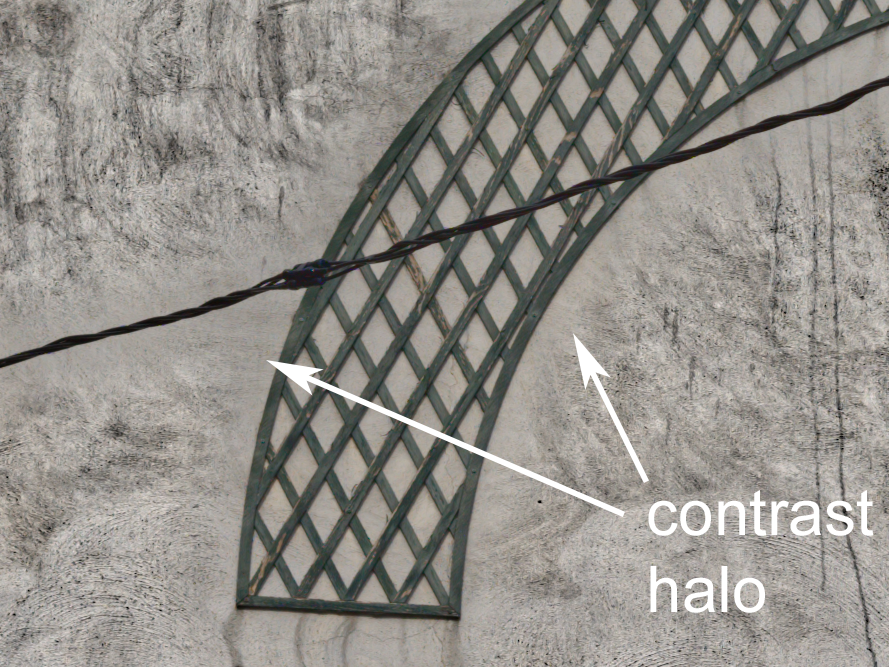} &
    \includegraphics[width=.238\linewidth]{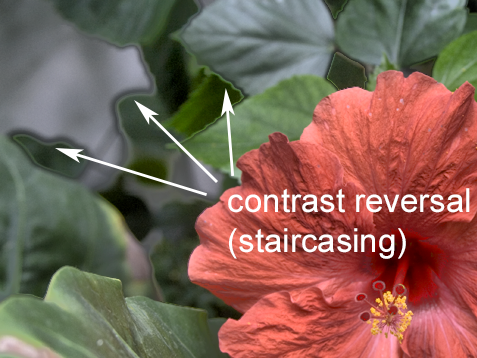} &
    \includegraphics[width=.238\linewidth]{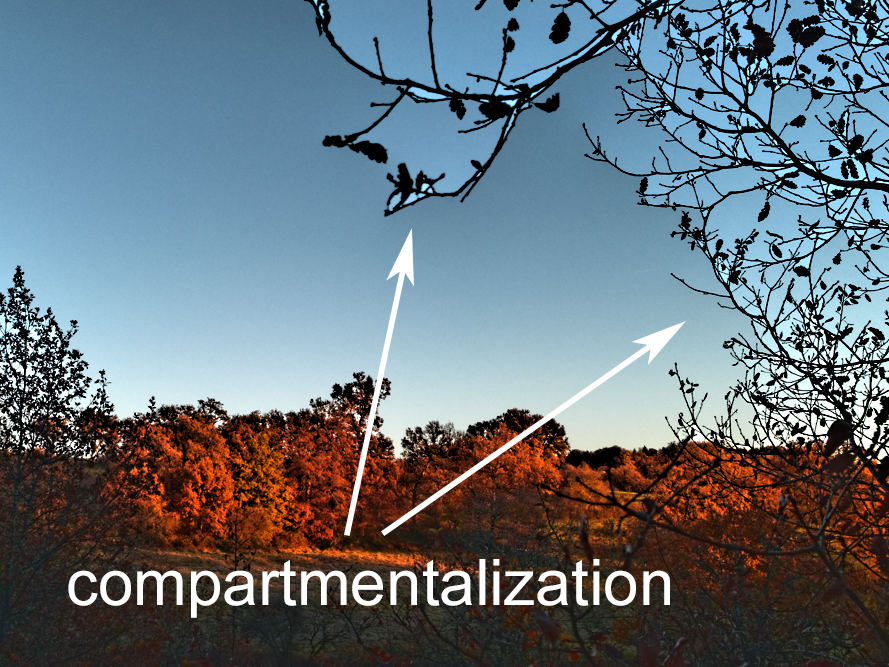} \\
    \textfg{(a) luminance halo} &
    \textfg{(b) contrast halo} &
    \textfg{(c) staircasing} &
    \textfg{(b) compartmentalization}
  \end{tabular}
  \caption{A panorama  of the image enhancement artifacts.
    In those contrast-enhanced images, the detail layer is amplified by
    $\text{\it enhance}(u) = \frac{1-\beta}{2}D + \beta\times \text{\it base}(u)
    + \alpha\times \text{\it detail}(u)$. In this formula, $D$ is the input
    dynamic range, $\beta=.75$ and $\alpha=3$ (except for the contrast halo
    where we used $\alpha=6$).
    The  luminance halo is the most  classic and well known artifact. The
    contrast halo attenuates texture near edges. The staircasing creates bands
    with inverted contrast often called the ``contrast reversal'' artifact.
    Compartmentalization effect breaks the unity of homogeneous regions.}
  \label{fig:teaser}
\end{figure*}

A key requirement of  such algorithms  is the absence of artifacts, that
originate from wrong attribution  of some base part  to the detail.
In the context of contrast enhancement, in which the dynamic of the base is
reduced and the  detail dynamics increased, minor errors may result in
conspicuous and unacceptable artifacts.  We  illustrate four of them in
Figure~\ref{fig:teaser}.
Motivated by the rapid development of digital photography, many decomposition
filters have been proposed since the early 2000s.
The presence of artifacts is acknowledged in most of them.
  It  often serves as argument for their comparison, but such arguments have so far remained partial and merely qualitative.

This study intends to provide experts with a clear identification of the
different types of artifacts in the base and detail decomposition filters
together with a methodology to measure each of them and to eventually give a reliable mark to each method.
To that aim, we shall first perform an analysis of the most prominent base+detail
decomposition filters and proceed to the identification of
their respective artifacts.
We then propose specific patterns, designed to stir each
targeted artifact.
Our mathematical definition of each artifact, coupled
with its dedicated pattern, yields a measure that can be associated with each edge-aware smoothing
filter.
 We are  led to address the delicate  problem of comparing fairly  the filters
 based on pattern measurements.  This requires a cross-calibration of base+detail filters so that they yield  the same amount of detail. Once the parameters of the filters have been fixed accordingly,  they can be
applied to  our  patterns and yield a mark for each filter and each artifact.
The last  problem addressed here is to find the adequate weights for  each artifact
measurement, taking into account that  some  of them like the  luminance halo
are  less annoying  than for  example  the staircase effect. To find the right
combination,  we  rely on annotations by experts.  In that  way we  transform
partial and subjective rejections by experts into an automatic global quality
measurement  for each  filter.   It  delivers a rank for  any base+detail
method and  actually leads us to rank objectively the state of the art methods.


In summary, we make the following contributions:
\begin{itemize}
  \item We draw up a list of artifacts likely to appear in any base and detail
    decomposition filter;
  \item We analyze and explain the identified artifacts, and propose for each of
    them a  pattern-measure pair to quantitatively evaluate its presence in
    any filter;
  \item We design a simple  cross-calibration method to set the filters'
    parameters, so that they can be compared fairly;
  \item We involve an expert evaluation (rather than non-specialist subjects)
    to fuse the independent measures of artifacts into a single
    meaningful score;
  \item We rank representative classic filters according to our measures, thus
    clarifying what filters are best suited for contrast enhancement.
\end{itemize}

The paper is organized as follows:
\begin{itemize}
  \item We first discuss in Section~\ref{sec:relatedwork} the broader topic of
    image quality assessment, that has related methodological issues, and the
    existing descriptions and measures of certain artifacts;
  \item  Section~\ref{sec:filterslist} reviews seven classic filters,  each
    being arguably the most representative filter in a different class of
    filters performing base+detail decomposition. We  analyze the guided filter,
    the weighted least squares filter, the local Laplacian filters, the total
    variation with $L^1$ data fidelity, the image smoothing via $L^0$ gradient
    minimization filter, and the domain transform.
  \item We define the artifacts in Section~\ref{sec:testpatterns}, give
    their mathematical definition and design patterns to measure them.
  \item Section~\ref{sec:params} addresses the  cross-calibration of filters
    previous  to their ranking.
  \item Section~\ref{sec:measures} performs the  ranking  and matches it to
    expert evaluations.
\end{itemize}

\section{Related work}
\label{sec:relatedwork}


Enhancement filters such as tone-mapping operators and local contrast
enhancement filters  are prone to artifacts. Indeed, the detail layer's dynamic
is expanded,  thus enhancing  any error in the base+detail decomposition, that
becomes a conspicuous artifact.
To the best of  our knowledge, no previous paper has proposed ranking filters
by measuring  their  artifacts.  In  the Reinhard \etal{} book on high dynamic
range imaging \cite{reinhard2010high}, one can read for example that
\begin{quote}
  Recently, some attempts of no-reference image contrast, sharpness, color
  saturation, and presence of noise evaluation have been reported; however,
  combining all these factors into a meaningful quality prediction is a
  difficult problem.
\end{quote}

Numerous perceptual studies have been carried out for high dynamic range
(HDR) imaging, to evaluate the  subjective  quality of  images obtained by  different tone-mapping operators (TMOs).
As stated by Drago \etal{} \cite{drago2003adaptive}:
\begin{quote}
  Essentially, tone mapping should provide drastic contrast reduction from scene
  values to displayable ranges while preserving the image details essential to
  appreciate the scene content.
\end{quote}

Eilertsen \etal{} \cite{eilertsen2013evaluation} classify the TMOs in three
categories: the \emph{visual system simulators} (VSS) that simulate the human
visual system, the \emph{scene reproduction} (SRP) operators that try to
reproduce as faithfully as possible the HDR scene, and the \emph{best subjective
  quality} (BSQ) operators that are designed to produce good-looking images
irrespective of the original content.  Technically, one can distinguish two
types of TMOs: the \emph{global} and the \emph{local} operators.
The global operators apply the same correction to pixels with the same
color, whereas in the local operators the correction depends on the local
content. Global TMOs are faster and generate fewer artifacts, but local TMOs
better preserve the local contrast. A base+detail decomposition therefore is
the core of most local TMOs  \cite{eilertsen2017comparative}.

A first set of studies assesses image quality based on
evaluations by non-expert subjects. This concerns most of the published
evaluation methods, which then can be divided in two categories. The first
category measures the similarity of a tone-mapped image to the original HDR
scene or to the HDR image displayed with an adapted screen
\cite{yoshida2005perceptual, ledda2005evaluation, cadik2006image,
  kuang2007evaluating, kuang2010evaluation}.
The second set of studies propose subjective image quality evaluation without
referring it to the original one
\cite{drago2002perceptual, drago2003perceptual, cadik2006image,
  kuang2007evaluating, kuang2004testing, kuang2005image, akyuz2007do,
  eilertsen2013evaluation}.

Another group of papers proposes objective measurements
for basic properties of tone-mapped images, such as color, contrast and
well-exposedness. Here again, the measures can be decomposed in two categories,
known under the name of full-reference
\cite{aydin2008dynamic, yeganeh2010objective, yeganeh2013objective,
  eilertsen2017comparative, mantiuk2011hdr}
and no-reference quality metrics \cite{aydin2015automated, mertens2007exposure,
  mertens2009exposure}.

For a detailed presentation of the cited papers, we refer to the
supplementary material provided with this paper.
For a more complete and general review of the quality assessment method we refer
to \cite{reinhard2010high, lavoue2015visual, wang2006modern, lin2011perceptual}.

These numerous studies for image quality assessment and their divergent or even
contradictory conclusions point to the  lack of universally accepted
definitions and  measurements of image quality.  All studies we mentioned are
motivated by the  emergence of more  and more effective TMOs, applied to the
richer content of an HDR image.  This combination has  introduced new degrees of
freedom in image rendering. The  new local TMOs are indeed incredibly flexible,
and can at will enhance locally the  image  and manage its colors.  Given these
new degrees of freedom, the reviewed studies primarily aim at establishing
quantitative aesthetic criteria to orient  TMO's and fix their parameters.   The
variety of image quality criteria indicates that they are subjective,
culture-dependent. This  is why  they must be calibrated by subjects.

Our goal here is more restricted and  focuses on the purely image  processing
side of the question.  We intend to  rank methods not by their final image
aesthetic quality, but by what they  are forbidden to  deliver.  In other words,
a ``good'' TMO should be able to create tasty or tasteless exaggerated images as
well. The  goal of image processing here  is rather to provide  the maximal
freedom to photographers and camera designers.

In short we have two separate problems. One is the aesthetic quality assessment
of images produced by TMOs  with well selected parameters. The other problem is to
evaluate how freely these parameters can be set without producing unacceptable
image quality flaws (artifacts!).

Indeed, as will be confirmed in the next section, the  main limitation to image
operators is the artifacts they produce. Defining a correct artifact measure
helps deciding which base+detail method has most degrees of  freedom. Thus,
this method will be  the best, as it can more freely adapt to any imaging
design.

\paragraph*{Organization}
This section is organized in two subsections.
The first is devoted to the literature on artifacts, their discovery and
measurement.  The second subsection reviews  how  papers have  addressed the
challenging  question of setting the  parameters of  each method in  a
comparison benchmark.

\subsection{Existing description and measures of artifacts}

While artifacts are the most frequent  reason  invoked to propose better algorithms,  no serious attempt to define quantitatively and measure those artifacts has been
proposed so far. The presence of the artifacts, on the other hand, is testified
in a number of papers.

In \cite{lavoue2015visual} for example, the presence of artifacts in the local
tone-mapping operator is invoked to explain why a number of studies
conclude that global operators are preferred over the local ones. When
\v{C}ad\'{i}k \etal{} \cite{cadik2008evaluation} included in their perceptual
study\footnotemark{} the notion of artifacts, it appeared that this attribute
participated to a large extent to the global perceived quality.

\footnotetext{Described in supplementary material, section related work,
    subsection perceptual studies.}

Artifacts are often invoked as a reason to propose new base+detail
decomposition filters. Furthermore, comparisons between
base+detail filters are generally made on some difficult images that generate artifacts;
so that comparing the filters amounts to comparing their artifacts. However, no
exhaustive list of the artifacts has been proposed yet.  This is problematic,  as often a new
method solves one artifact while introducing another one, which is later
  uncovered by  another  paper.
This is the case for example for the weighted least squares filter (\ffwls{}) \cite{farbman2008edge}.

In this paper, Farbman \etal{} consider the problem of multi-scale
  detail enhancement. They show that former existing schemes produce artifacts: either
a \emph{staircase effect} in the scheme of Fattal \etal{}
\cite{fattal2007multiscale}, or a \emph{luminance halo} in the scheme of Chen
\etal{} \cite{chen2007real}. They then show that their proposed filter does
not create any of these artifacts.
As we shall see, their filter actually introduced a strong
\emph{compartmentalization} artifact, which would later be pointed out, for
example in \cite{he2010guided}.

Chronologically, the next successful edge-aware smoothing filter is the guided
filter (\ffgf{}) \cite{he2010guided}.  In this paper, \ffgf{} is compared
against the bilateral filter (\ffbf{}) \cite{tomasi1998bilateral}. The  authors argue that their filter avoids \emph{staircasing},
and an ``intensity shift'' artifact in \ffwls{},
which corresponds to what we shall call \emph{compartmentalization}.
Yet, \ffgf{} was actually introducing a new artifact, the \emph{contrast halo}.

In 2011, Gastal \etal{} proposed a fast filter called \emph{domain transform}
(\ffdt{}) \cite{gastal2011domain}. Its compartmentalization was not compared to that of
 previous filters. Yet DT produces it the most, as we shall see in
Section~\ref{sec:measures}.
As for the contrast halo artifact of \ffgf{}, it was also precisely described in
\cite{lu2012cross}.
In short, the artifacts are known, sometimes under different names, and serve as reference for the comparisons. Yet, with the exception of  the \emph{luminance halo} there is no proposed measurement for each, and still less a joint measurement.

Of the four artifacts covered in this paper, the most known and most discussed
one is unquestionably the luminance halo, often simply called \emph{halo}.
It has been studied  by Trentacoste \etal{} \cite{trentacoste2012unsharp} in
2012, where the authors analyzed its perception and wondered if it should be
viewed as an enhancement or as an artifact. Both interpretations indeed coexist.
The authors show that the decision between both depends on the halo's
amplitude and width.
In this paper, the authors propose a contrast enhancement method based on
countershading. They conducted a perceptual study to determine the parameters of
their method, so that ``local contrast can be introduced by [their] operator
without becoming objectionable''.
They modeled the objectionable threshold on the magnitude of the countershading
profile in function of the standard-deviation of the Gaussian filter used to
generate it. This unfortunately is specific to the Gaussian filter and does not
allow to measure the luminance halo generated with other methods.

Li, Sharan and Adelson in 2005 \cite{li2005compressing} largely discuss the halo
artifact, as they method is based on the multi-scale manipulation of Laplacian
coefficients. They succeed in masking the luminance halo by choosing adequate
weights; yet here again no attempt was made to measure it.

Jang \etal{} \cite{jang2011local} proposed a method to automatically adjust the
parameters of the multi-scale retinex algorithm, so as to optimize the
perceptual quality of the contrast-enhanced image.  A measure of the contrast was
obtained using local standard-deviations. To set the weight and width of the
Gaussian filters, they proposed a measure to evaluate the presence of halos on test images.
The halo artifacts were evaluated based on the maximum difference between each
pixel intensity on one side of the edge and the average intensity on the same
side. The maximum intensity difference corresponding to each side of the edge is
computed as
\begin{equation}
h_s = \max_{x \in \Omega_s}\left( | I(x) - \bar{I} | \right),
\ s \in \{\text{left}, \text{right}\},
\end{equation}
where $\Omega_s$ represents the pixels on the left or right side of the edge and
$\bar{I}$ represents the mean value. Finally, the overall halo artifact measure
is obtained based on the averaged maximum intensity differences for each edge
side:
\begin{equation}
H = \frac{1}{2} ( h_\text{left} + h_\text{right} ).
\end{equation}
We shall use a similar measurement on our evaluation pattern.

\subsection{Setting the parameters}
\label{subsec:relatedwork:params}

In virtually every quality assessment paper, the authors have to decide how the
parameters of the different tone-mapping operators will be fixed. Most of the
time, they rely on the parameters given by the authors, but while this seems
correct for TMOs with the same \emph{intent}\footnotemark{}, it does not make sense when the
parameters were set with a different objective in mind.
We refer to \cite[Section 7.2.3]{eilertsen2016evaluation} for a review of
the different ways used in TMO evaluation for the parameter settings. In this
book section, the authors consider three groups: 1)~use of default parameters;
2)~tuned by experts; 3)~pilot study.

\footnotetext{See supplementary material, section related work,
    subsection perceptual studies.}

In their evaluation of TMOs for HDR-video in 2013
\cite{eilertsen2013evaluation}, Eilertsen \etal{} described an original way of
setting the parameters of the tested operators.
They asked four experts to tune the parameters so that they
produce good results in three sequences, using Powell's conjugate direction
method, where at least two full iterations were completed. Finally the
parameters obtained with the different experts were averaged and then used in
two subjective evaluation experiments\footnotemark{}.
This process of manual tweaking can be helped by \emph{perceptually linearizing}
the parameter space \cite{eilertsen2014perceptually}.

\footnotetext{Described in supplementary material.}

All such methods rely on subjective evaluations of the final tone-mapped images.
Our problem is different, for two reasons. First, there is no
reference for the detail image; second, it is unclear what the best-looking
detail layer would be, since this type of image is quite unnatural.
In fact, we are not interested here in the subjective quality of the
  detail (nor of the enhanced images), but in the artifacts they may introduce
  for a comparable enhancement of the contrast.
  As said earlier, our concern is to evaluate the range in which a filter can be
  used without introducing objectionable artifacts. This amounts to evaluating
  the strength of artifacts when the filters produce the same amount of
  detail.
In Section~\ref{sec:params} we describe our method, which sets the parameters of each filter so that the average $L^2$-norm of
  the detail layer is equal for each
  filter.

\section{A list  of representative filters}
\label{sec:filterslist}


\subsection{Criteria for the  choice of the  filters}

We shall restrict ourselves to seven filters. Each one  is an acknowledged
representative of a  wide  class  of filters. We list  them together with their abbreviations.
\begin{enumerate}
  \item the bilateral filter (\ffbf{}),
  \item  the guided filter (\ffgf{}),
  \item  the weighted least squares filter (\ffwls{}),
  \item  the local Laplacian filters (\ffllf{}),
  \item the total variation with $L^1$ data fidelity (\fftv{}),
  \item the image smoothing via $L^0$ gradient minimization (\fflois{}),
  \item the domain transform (\ffdt{}).
\end{enumerate}
The classes of algorithms under consideration are quite different.  The
bilateral belongs to the wide class of  neighborhood filters that perform a
nonlinear local  convolution.   The  guided filter belongs to the  class of
anisotropic filters.  The  weighted least squares filter derives from  a
variational edge aware model,  the local Laplacian filter is wavelet based and inherently
multiscale. The  total  variation is a functional analysis model, the $L^0$ and
$L^1$ are sparsity models, and the  domain  transform has an underlying
image+color  manifold model.

These filters  are representatives of  different image structure  models, which lead to a different notion  of detail  and  therefore to different artifacts  when the detail is  enhanced.

The selected filters and their acronyms are listed in Table~\ref{table:params},
along  with their parameters. The methodology we use to set the parameters is
described in Section~\ref{sec:params}.  We shall now briefly present each filter
with its definition and the artifacts it introduces\footnotemark{}.

\footnotetext{More filters are presented in supplementary material: those
  that we do not consider because they would be redundant or are too
    complex, and others that directly modify the local contrast without
    base+detail decomposition.}

\begin{figure*} 
  \centering
  \setlength{\tabcolsep}{.01\linewidth} 
  \begin{tabular}{@{}cccc@{}}
    \includegraphics[width=.235\linewidth]{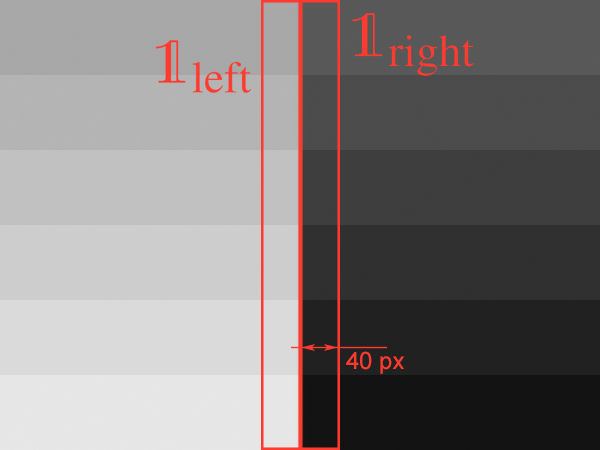} &
    \includegraphics[width=.235\linewidth]{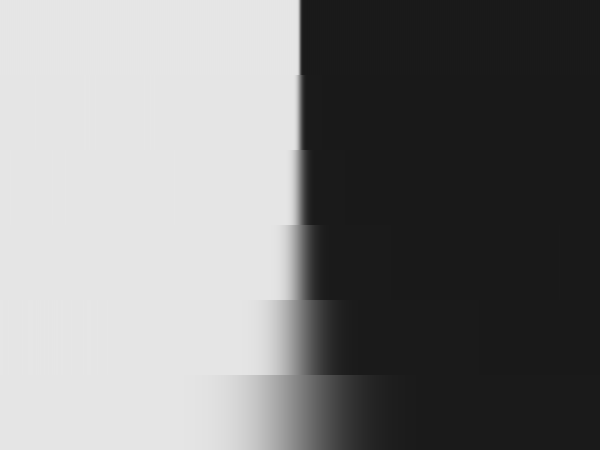} &
    \includegraphics[width=.235\linewidth]{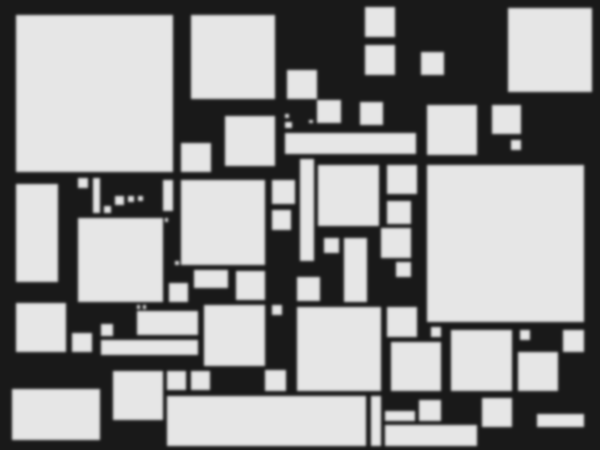} &
    \includegraphics[width=.235\linewidth]{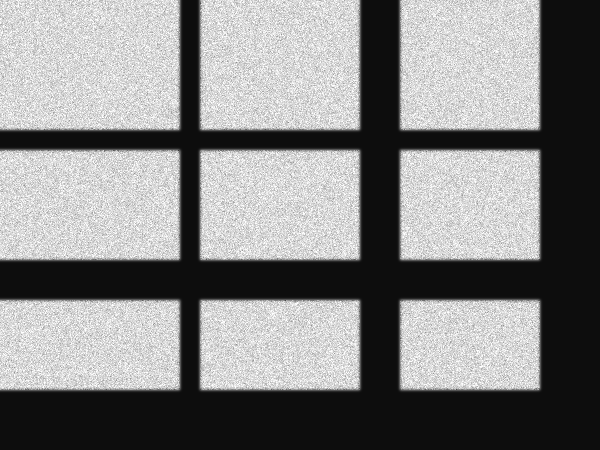} \\
    \textfg{(a) luminance halo} &
    \textfg{(b) staircasing} &
    \textfg{(c) compartmentalization} &
    \textfg{(d) contrast halo}
  \end{tabular}
  \caption{Test patterns for artifacts measurements. The first two patterns (a)
    and (b) are in fact a series of 6 patterns composed of one single vertical
    edge, as explained in section~\ref{subsec:patterns:lumhalo}. We display here
    for these two patterns a stack of horizontal bands taken from each of the
    input.}
  \label{fig:patterns}
\end{figure*} 

\subsection{Presentation of the seven filters}
\label{subsec:filterslist:definitions}

\subsubsection{The  bilateral filter} 

First intended for denoising, this filter appeared in
1983 with Yaroslavsky \cite{yaroslavsky2012digital} and Lee
\cite{lee1983digital}. The variant using two Gaussian functions was proposed by
Smith and Brady who called it ``SUSAN'' (1995) \cite{smith1997susan}. It was
discovered again by Tomasi and Manduchi in 1998 \cite{tomasi1998bilateral} who named it ``bilateral filter''.
It is defined by
\begin{equation} \label{equ:bilateral}
  \imm(\x) = \frac{1}{C(\x)} \sum_{\y \in \Omega}
  e^{\frac{\|\x-\y\|^2}{2\sigma_s^2}}
  e^{\frac{|\im(\y)-\im(\x)|^2}{2\sigma_r^2}} \im(\y),
\end{equation}
where $\im$ and $\imm$ are the input and filtered images, respectively; $\x$ and
$\y$ are pixels positions and $\Omega$ a window. $C$ is the normalization
factor.
This filters has many variants, for example the joint bilateral filter and the
unnormalized bilateral filter to name a few. Numerous fast approximations have
been proposed. We refer to the Paris \etal{} book \cite{paris2009bilateral} for
a thorough review.
We shall consider in this paper the most representative of its fast
approximations, namely, the bilateral grid \cite{paris2006fast, chen2007real}.
To avoid any artifact due to the approximation, the range subsampling will be
set to be very fine, \ie{}, $S=64$ with $S$ the number of slices, considering
that the parameter $\sigma_r \simeq 0.1$ (image dynamic in $[0,1]$).


\subsubsection{The domain transform \texorpdfstring{\cite{gastal2011domain}}{gastal2011domain}} 

The idea of this fast filter  is  to  make  a nonlinear monotone domain
transform on each image line so  that the bilateral filter applied  to each line
boils down to  a convolution.  The  image is  then alternately filtered in  line
and in column to  emulate  a 2D bilateral filter. On each line the image coordinate $z$ is redefined as a strictly increasing 1D signal  $ct(z)$ by
\begin{equation} \label{equ:domaintrans}
  ct(z) = \int_0^z 1 +
  \frac{\sigma_s}{\sigma_r} \sum_{k=1}^{c} |\im_{x,k}(x)| dx,
\end{equation}
where $\sigma_r$, $\sigma_s$ are the filter's parameters and $\im_{x,k}$ is the
derivative of the input image $\im$ for channel $k$ and pixel $x$. This amounts
to defining the new distance between image points $0$ and $z$ as the
geodesic distance between $(0, \im(0))$  and $(z,  \im(z))$ in the  image
graph.  As a result, pixels with distant intensities fall apart. Hence, a
Gaussian filter applied on the transform domain averages them but little. 

\subsubsection{The guided filter \texorpdfstring{\cite{he2010guided}}{he2010guided}} 

This  filter, ubiquitous in image processing
since its publication in 2010,  is defined by
\begin{equation} \label{equ:gf}
  \ffgf\{\im\}(\x) = \bar{a}(\x) \imm(\x) + \bar{b}(\x),
\end{equation}
where the bar $\bar{.}$ means a mean in the neighborhood defined by a square
window $\Omega$ and the linear coefficients $a$ and $b$ are obtained by
minimizing the cost function
\begin{equation} \label{equ:gfcostfunc}
  E\big( a(\y), b(\y) \big) = \sum_{\x \in \Omega(\y)}
  \Big( \big( a(\y) \imm(\x) + b(\y) - \im(\x) \big)^2
  + \epsilon a(\y)^2 \Big),
\end{equation}
where $\epsilon$ is the smoothing parameter, $\im(\x)$ the original image and
$\imm(\x)$ the guide.


\subsubsection{The  weighted least squares filter (\ffwls{})
  \texorpdfstring{\cite{farbman2008edge}}{farbman2008edge}} 

Given an input image $\im$, this energy-based filter seeks an output $\imm$,
\begin{quote}
  which, on the one hand, is as close as possible to $\im$, and, on the  other hand as smooth as possible everywhere, except across significant gradients
  in $\im$.
\end{quote}
This translates into finding an image $v$ minimizing
\begin{equation} \label{equ:wls}
  \sum_{\x} \Bigg(
  \big(\imm(\x) - \im(\x)\big)^2 + \lambda \!\!\sum_{z\in\{x,y\}}\!\!
  a_{z}(\im,\x) \left( \frac{\partial \imm}{\partial z} \right)^2\!\!(\x)
  \Bigg),
\end{equation}
where $a_x(\im,\x)$ and $a_y(\im,\x)$ are smoothness weights defined as
\begin{equation} \label{equ:wlscoefs}
  a_{z}(\im,\x)
  = \left( \left| \frac{\partial \ell }{\partial z}(\x) \right|^\alpha
  + \epsilon \right)^{-1},
\end{equation}
where $\ell$ is the \emph{log-luminance} channel of $\im$.

\subsubsection{The local Laplacian filter
  \texorpdfstring{\cite{paris2015local}}{paris2015loca}} 

This filter is considered as the highest  quality  base and detail
decomposition, to the  price of a higher complexity.
Although inspired from the bilateral filter, it presents almost no
  staircase effect.
This filter is versatile and can be used for a wide variety of
contrast manipulations tasks, ranging from edge-aware smoothing to local
contrast enhancement with dynamic reduction. It directly  computes
the Laplacian pyramid of the output image; a final operation collapses
the pyramid and builds the filtered image.  Each Laplacian coefficient is
computed independently using a dedicated \emph{remapping function}, which shape
is chosen in function of the application.
The fast version (\fffllf{}) uses the Durand-Dorsey \cite{durand2002fast}
slicing strategy. It greatly speeds up the execution by computing only a
reduced number of remapped images.

The filter can be summarized  in the next  formula,
\begin{equation} \label{equ:llf}
  \lpyr\{\imm,l\}(\x) = \sum_{i=1}^{S}
  A_i(l,\x) \lpyr\{\im'_i,l\}\!(\x),
\end{equation}
where $\lpyr\{\imm\}$ is the Laplacian pyramid of the output image $\imm$, $l$ is
the scale, $\x$ is a pixel, $\im$ is the input image and $\im'_i$ the input 
remapped for the sample $i \in \{1,2,\ldots,S\}$. $A_i$ is an weight map of the
same size as $\lpyr\{\im'_i,l\}$.
The final image $\imm$ is obtained by \emph{collapsing} the constructed pyramid
\cite{burt1983laplacian}.

\subsubsection{Total variation} 

The total variation model assumes that the  total variation of  the  base is bounded while the detail would be  highly oscillatory and therefore only   integrable. This leads to a
TV+$L^1$ minimization  which is classically performed by the Chambolle-Pock method \cite{chambolle2011first},  as implemented in \cite{guen2014cartoon}.
This filter finds $v$  minimizing
\begin{equation} \label{equ:tvlone}
   \|\im - \imm\|_1 + \lambda \|\nabla\imm\|_1 ,
\end{equation}
where $\lambda$ is the smoothing parameter, $\im$ the input image, $\imm$ the
smoothed one (the base layer) and $\im - \imm$ the  detail layer.  This  filter was designed for the cartoon+texture decomposition   introduced by Meyer \cite{meyer2001oscillating}.

\subsubsection{%
  Image smoothing via \texorpdfstring{$L^0$}{L0} gradient minimization} 

This successful edge-aware smoothing filter   proposed  by
Xu \etal{}  \cite{xu2011image} is based on the minimization of the $L^0$~norm of the gradients of
the output image $\imm$. We shall call it \fflois{} in the following.  It
consists in minimizing the number of non-zero gradients while ensuring that the
output stays close enough to the original image with a quadratic data
attachment term. Formally, the method seeks an image $\imm$ that minimizes
\begin{equation} \label{equ:lois}
  \|\im - \imm\|_2^2 + \lambda C(\imm) ,
\end{equation}
where $C()$ counts the number of pixels whose gradient is not zero, and
$\lambda$ is a parameter that controls the amount of smoothing.  The second term  requires the  base's gradient to be sparse.


\section{The  main  artifacts and their test patterns}
\label{sec:testpatterns}

In this section we present the measures designed to quantitatively evaluate the
four canonical artifacts of  edge-aware filters. Each measure is associated with
a test-pattern specifically designed to detect and measure one of the artifacts.
This evaluation is limited to the artifacts  representing the main
impediments in the base+detail filters.
They are the \emph{luminance halo}, the \emph{staircasing}, the
\emph{compartmentalization} and the \emph{contrast halo}.
The corresponding test-patterns are displayed in
Figure~\ref{fig:patterns}.  For  each artifact,  we shall point out the
filters that produce them most.

\subsection{The luminance halo}
\label{subsec:patterns:lumhalo}

The luminance halo is the most common artifact of contrast enhancement filters,
already present in high pass filters.  It arises when a contrasted edge has been
smoothed, even slightly, while it should have been preserved in the  base.

To measure the luminance halo, it is  enough to build a test-pattern containing
flat regions separated  by straight edges of all amplitudes.  The measure then
simply quantifies the distance between the filtered image and the input one.

\begin{figure} 
  \centering
  \setlength{\tabcolsep}{.0175\linewidth}
  \begin{tabular}{@{}ccc@{}}
    \includegraphics[width=.31\linewidth]{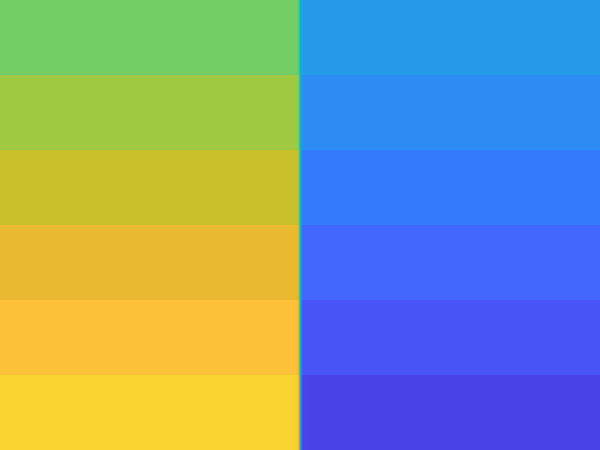} &
    \includegraphics[width=.31\linewidth]{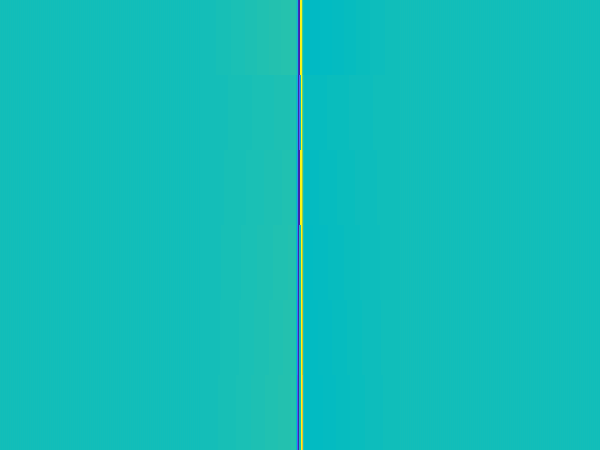} &
    \includegraphics[width=.31\linewidth]{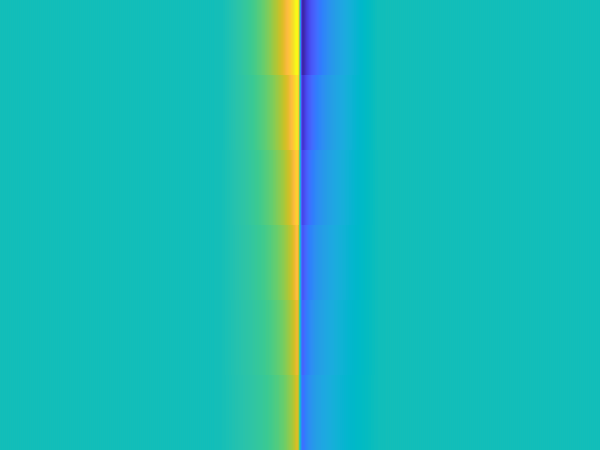} \\
    \textfg{(a) input} &
    \textfg{(b) detail with \fffbf{}} &
    \textfg{(c) detail with \ffgf{}}
  \end{tabular}
  \caption[Luminance halo test pattern and result with associated worst filter]{
    Luminance halo artifacts with \ffgf{}.  This triptych presents bands
      of the input patterns in (a), bands of the detail layers obtained with
      \fffbf{}, which has no luminance halo (b), and bands of the detail layers
      obtained with \ffgf{} which has the highest halo score
      (c).
      All of these images are represented with false colors.  Dynamic range is
      $[0,1]$ for~(a) and $[-0.04,+0.04]$ in~(b) and~(c).}
  \label{fig:patterns:worstlumhalo}
\end{figure} 

\paragraph{Test-pattern}
For this pattern, as well as for the staircase effect test pattern
  (described in~\ref{subsec:patterns:staircase}~(a)) we actually use a series of six patterns. This is because we have two
  contradictory needs. First, we need to test different edge heights. Yet we
  also need to dispose of patterns with only one edge. Indeed, this ensures that the deformation
  caused by the filter (the artifacts!) does not comes from another nearby edge.
 The six patterns have a centered vertical
  edge with different heights (we used $\{ 0.3,$ $0.4,$ $0.5,$ $0.6,$ $0.7,$
  $0.8 \}$ for this pattern).  Each pattern is filtered with the
  tested filter, then a horizontal band (with the same width but only one
  6$^\text{th}$ of the initial height) is extracted in the middle of the
  pattern. The 6 bands are stacked together to create a single output image.
  This composite base layer is then used for the halo measurement. This explains why the
  detail layers only display horizontal luminance halos (for example in
  Figure~\ref{fig:patterns:worstlumhalo}~(c)).

\paragraph{The halo measure}
\label{par:patterns:lumhalomeasure}
Based upon this test-pattern, the halo measure  is defined by

\begin{equation} \label{equ:measure:lumhalo}
  \begin{split}
    \mathcal{H}(\im_0,\im_1) =
    \sum_{\x \in \Omega} \mathds{1}_{\text{left}}(\x)
    \left(\left( \im_0(\x) - \im_1(\x) \right)^{+} \right)^2 \\
    + \sum_{\x \in \Omega} \mathds{1}_{\text{right}}(\x)
  \left(\left( \im_1(\x) - \im_0(\x) \right)^{+} \right)^2,
  \end{split}
\end{equation}%
where $\x$ is a pixel, $\im_0$ is the input image, $\im_1$ is the filtered image
and $\mathds{1}_\text{left}$, $\mathds{1}_\text{right}$ are indicator functions.
These functions equal $1$ in rectangles of width $40$ pixels (the standard-deviation of the
tested filters) and are placed on the left and right side of the vertical edge
respectively. The rectangles are displayed in Figure~\ref{fig:patterns}~(a).
We denote by $\Omega$ the image domain and $(.)^{+}$ denotes
the positive part.
Because a few strong differences are more annoying than numerous small ones, we
square the positive part in Equation~\eqref{equ:measure:lumhalo}.

\paragraph{Which  filter performs worst}
The worst filter for the luminance halo artifact is the guided filter.
 Figure~\ref{fig:patterns:worstlumhalo} shows the detail layer given by this
filter for our test-pattern, and compare it to the detail layer obtained
with the fast bilateral filter, which creates the least luminance halo.
The parameters of both filters are given in Table~\ref{table:params}.

\subsection{The staircase effect}
\label{subsec:patterns:staircase}

The staircase effect is typically present in the bilateral filters; it is
sometimes referred to as an edge sharpening effect.
Since the sign of the edge's gradient in the detail layer is
  opposite to its sign in the input image, enhancing the detail causes ``gradient reversal''
  in the final detail-enhanced image, which is yet another way to refer to this artifact.
Numerous correction schemes
have been proposed \cite{paris2009bilateral} but often fail at effectively correcting this artifact; even the
bilateral filter with regression \cite{buades2006staircasing} that has been
proposed specifically to solve this artifact can present it in some conditions
such as a large spatial support.

\begin{figure} 
  \centering
  \setlength{\tabcolsep}{.0175\linewidth}
  \begin{tabular}{@{}ccc@{}}
    \includegraphics[width=.31\linewidth]{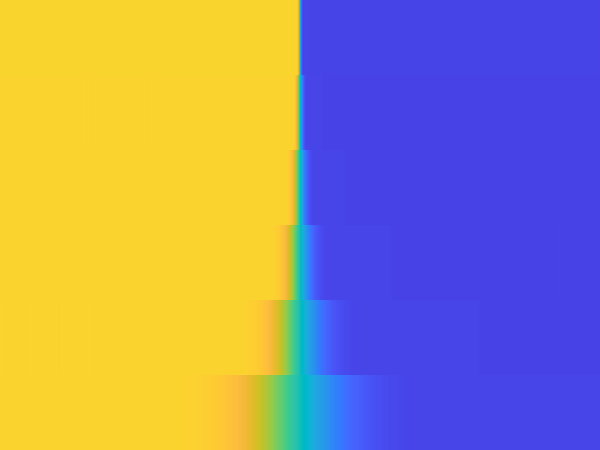} &
    \includegraphics[width=.31\linewidth]{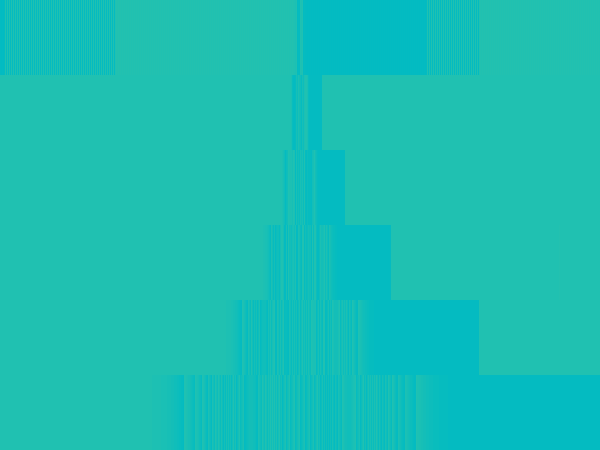} &
    \includegraphics[width=.31\linewidth]{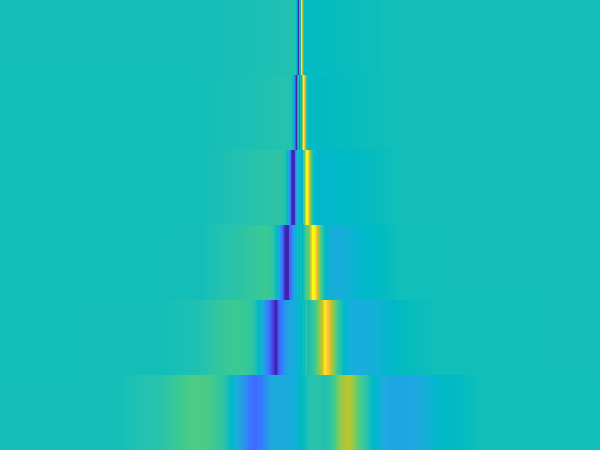} \\
    \textfg{(a) input test pattern} &
    \textfg{(b) detail with \fftv{}\,\footnotemark{}} &
    \textfg{(c) detail with \fffbf{}}
  \end{tabular}
  \caption[Staircase effect test pattern and result with its worst filter]{
    Staircasing test pattern (a). Detail layer with \fftv{} which gives
      the smallest score and with \fffbf{} which gives the highest.
    The input test pattern~(a) and its decomposition are represented with false
    colors. Dynamic range is $[0,1]$ for~(a) and $[-0.1,+0.1]$ in~(b) and~(c).}
  \label{fig:patterns:worststaircase}
\end{figure} 

\footnotetext{\label{footnote:tvstranglelines}%
  Horizontal oscillations with amplitude 1 appear in \fftv{} results.
  This is due to the conversion from double to unsigned 8-bits integer needed in
  the implementation we used. This is not due to the algorithm itself, and has a negligible influence on the measures.}

\paragraph{The staircasing test-pattern}
Like in the luminance halo pattern which construction is described in
section~\ref{subsec:patterns:lumhalo}, we use a set of six images to measure the
staircase effect. This allows us to filter images which contain only one
(smooth) edge but to test several edge widths.
Each of the six input pattern has a contrasted and blurred vertical edge. The
patterns are generated using a step edge convolved with a Gaussian kernel in the
Fourier domain using the standard-deviation $\sigma = \{ 0.7,$ $1.4,$ $2.8,$
$5.6,$ $11.2,$ $22.4 \}$.
Similarly to the luminance halo, we display a preview of the input patterns by
piling up horizontal bands into a single image shown in
Figure~\ref{fig:patterns}~(b).
The same process is used after filtering with a tested filter
so as to create a single output containing the six blurred edges. This output is
then used for the measurement.  This construction explains why the merged filtered image shows no interaction
between the bands; see Figure~\ref{fig:patterns:worststaircase}~(b)
and~(c).

\paragraph{The staircase measure}
Using the test-pattern we just described, we measure the edge reinforcement in
the six bands simultaneously using

\begin{equation} \label{equ:measure:staircase}
  \begin{split}
    \mathcal{S}(\im_0,\im_1) =
    \sum_{\x \in \Omega}  \mathds{1}_\text{left}(\x)
    \left( \left( \im_1(\x) - \im_0(\x) \right)^{+} \right)^2 \\
    + \sum_{\x \in \Omega} \mathds{1}_\text{right}(\x)
    \left( \left( \im_0(\x) - \im_1(\x) \right)^{+} \right)^2,
  \end{split}
\end{equation}
where the indicator functions $\mathds{1}_\text{left}$ and
$\mathds{1}_\text{right}$ are the same as for the luminance halo. They are
displayed in Figure~\ref{fig:patterns}~(a) and presented in
Section~\ref{subsec:patterns:lumhalo}.

\paragraph{Worst filters}
The worst filters are the bilateral filter and, without surprise, the
$L^0$ gradient minimization filter.
We display in Figure~\ref{fig:patterns:worststaircase}~(c) the detail layer
obtained with \fffbf{}.
On the detail image, the artifact appears as blue
bands on the left side of the edge and yellow bands on the right side. The
increasing width of the edge allows to determine the scale at which the artifact appears.
The worst case is always the finest edge, but the attenuation when the edge width
increases depends on the filter.
As control, we show the detail layer obtained with \fftv{},
  which has no staircase effect\footnote{The detail layers obtained with the different filters are observable in
  the supplementary materials.}.

\begin{figure} 
  \centering
  \setlength{\tabcolsep}{.0175\linewidth}
  \begin{tabular}{@{}ccc@{}}
    \includegraphics[width=.31\linewidth]{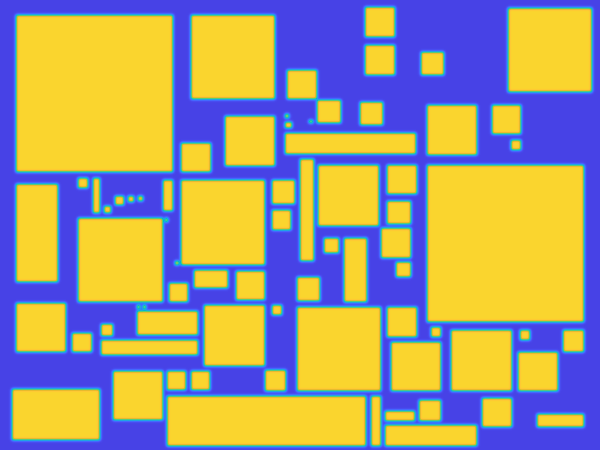} &
    \includegraphics[width=.31\linewidth]{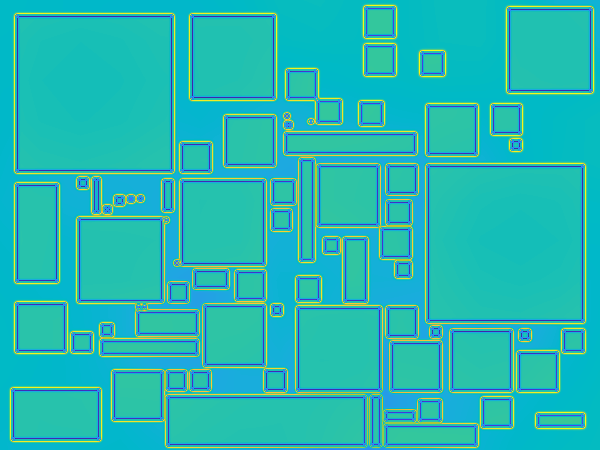} &
    \includegraphics[width=.31\linewidth]{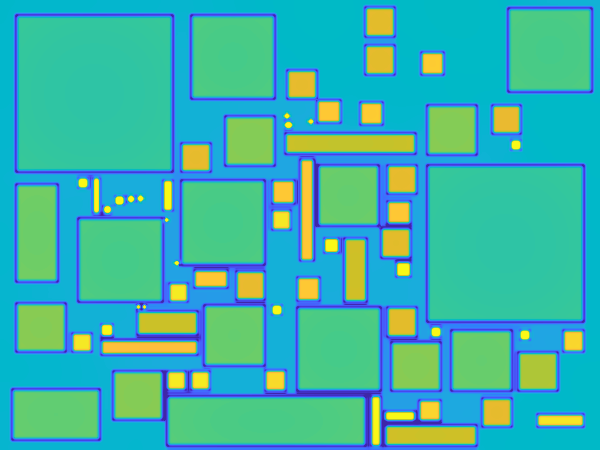} \\
    \textfg{(a) input test pattern} &
    \textfg{(b) detail with \fffbf{}} &
    \textfg{(c) detail with \ffwls{}}
  \end{tabular}
  \caption[Compartmentalization effect test pattern and result with its worst
    filter]{%
      This triptych shows the compartmentalization test pattern~(a),
        the detail layer obtained with the lowest score filter \fffbf{}~(b), and
        with one of the highest score filters, \ffwls{}~(c).
        All images are represented with false colors. Dynamic range is $[0,1]$
        for~(a) and $[-0.1,+0.1]$ in~(b) and~(c).}
  \label{fig:patterns:worstcompart}
\vspace*{\floatsep}
  \begin{tabular}{cc}
    \includegraphics[width=.667\linewidth]{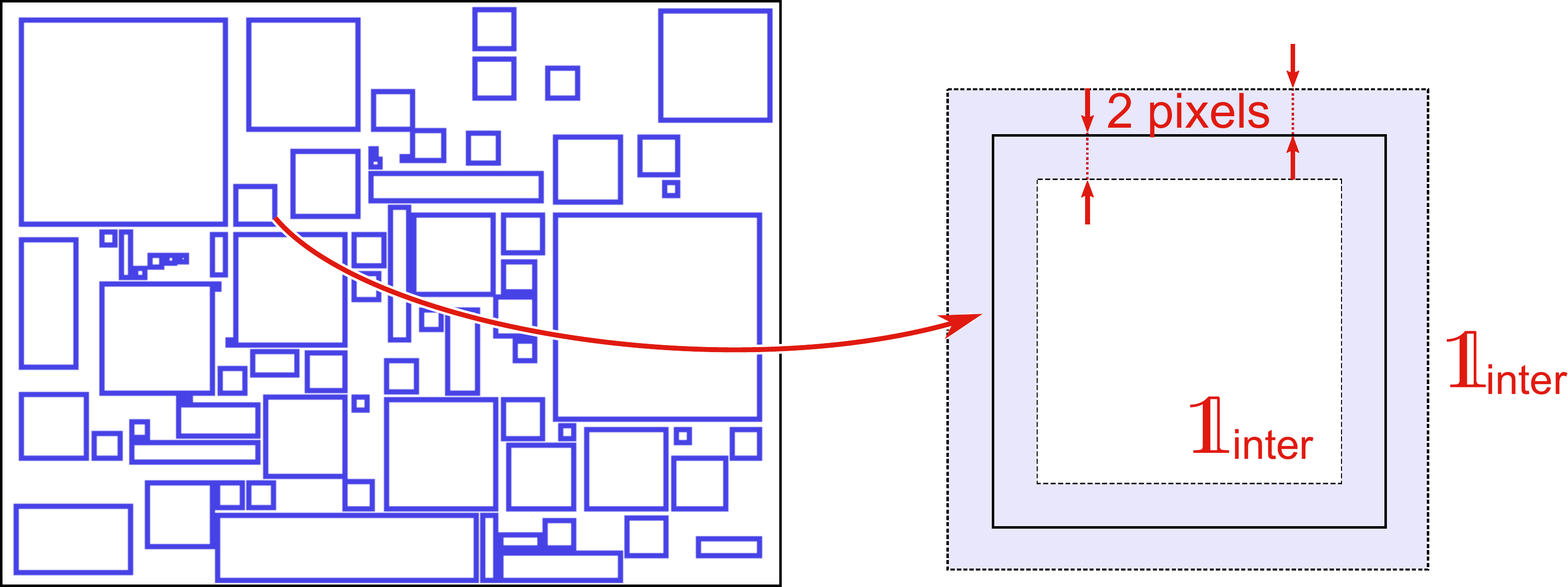} &
  \end{tabular}
  \caption[Masks used in the compartmentalization measure]{
    Illustration of the mask used in the compartmentalization measure: in order
    to quantify the alterations of the bright areas inside the shapes and
      the alteration of the dark interstices, we measure the variance of the
      detail layer using the mask $\mathds{1}_\text{inter}$ displayed in white
      in the figure. A two pixels wide area (the blue regions) is excluded on the
      inner and outer borders of the shapes, because of the slight smoothing we
    applied to avoid aliasing.}
  \label{fig:scheme:compart}
\end{figure} 

\subsection{The compartmentalization effect}
\label{subsec:patterns:compart}

The compartmentalization effect arises when a constant color region (typically a
wall, or the sky) is compartmented into pieces with variable size by the
superposition  of a  grid, or of tree branches,  etc.
The artifact consists in an intensity shift of  the  detail in constant regions; its magnitude
depends on the area of the region. The smaller the region, the stronger the
artifact.  The total variation~\cite{rudin1992nonlinear, chan2005image} presents
a drastic compartmentalization:  it removes local extrema with small
area from the  base and puts them fully in  the  detail, while  the  larger
extremal regions are  left in the base.

\paragraph{The compartmentalization test-pattern}
The compartmentalization test-pattern is made of bright squares and rectangles
of different areas disposed on a dark background (Figure~\ref{fig:patterns}).   This image is
slightly smoothed to avoid aliasing.

\paragraph{The compartmentalization measure}
Using the test-pattern described above, the measure is  defined by

\begin{equation} \label{equ:measure:compart}
  \mathcal{P}(\im_0,\im_1) =
  \mathrm{var}\left\{\left(\im_0-\im_1\right) \mathds{1}_{\text{inter}}\right\}
\end{equation}
where $\mathds{1}_{\text{inter}}$ is a mask corresponding to the image deprived
of a $2$-pixels wide band along the inner and outer borders of the shapes. This
excludes pixels influenced by the edge. This mask appears in white in the
illustration of Figure~\ref{fig:scheme:compart}. The removed pixels in the
squares are shown in blue.

\paragraph{Filters with worst compartmentalization}
A filter with spectacular compartmentalization artifacts is \ffwls{}.
Figure~\ref{fig:patterns:worstcompart} displays the result
obtained with this filter using the parameters set in
Section~\ref{sec:params} and compares it to the detail layer obtained with
  \fffbf{}, which does not have compartmentalization in~(b).
  One can recognize the staircase effect in image~(b) but the edges are not
  taken into account in the measure. Thus they do not influence the
  compartmentalization score.
With the false colors used, the bright regions appear in yellow and the dark
ones in blue; in the detail layer in~(c), one can clearly see that \ffwls{}
``lights up'' the small shapes.
The small interstices between the rectangles are  affected by the
  compartmentalization effect too; they appear as darkened zones in~(c).
  Our measure  takes it into account.
The highest score for the compartmentalization, however, is not obtained by
\ffwls{} but by the domain transform filter%
\footnote{The detail layers obtained with the different filters are observable in
  the supplementary material.}.
Yet the compartmentalization present
in \ffdt{} is related to its luminance halo artifact; in fact edges are not well
preserved by this filter, which causes the small elements to be smoothed out
even if their contrast is high. Thus, all shapes in the test-pattern are
affected.  For filters like \ffwls{} however, there is a distinct separation
between shapes that are affected (those that are lit in the detail layer)
and those that are not.

\subsection{Contrast halo}
\label{subsec:patterns:conthalo}

The contrast halo appears when  regions containing details and close to edges
are not filtered. This artifact is typical of the guided filter.

\begin{figure} 
  \centering
  \setlength{\tabcolsep}{.0175\linewidth}
  \begin{tabular}{@{}ccc@{}}
    \includegraphics[width=.31\linewidth]{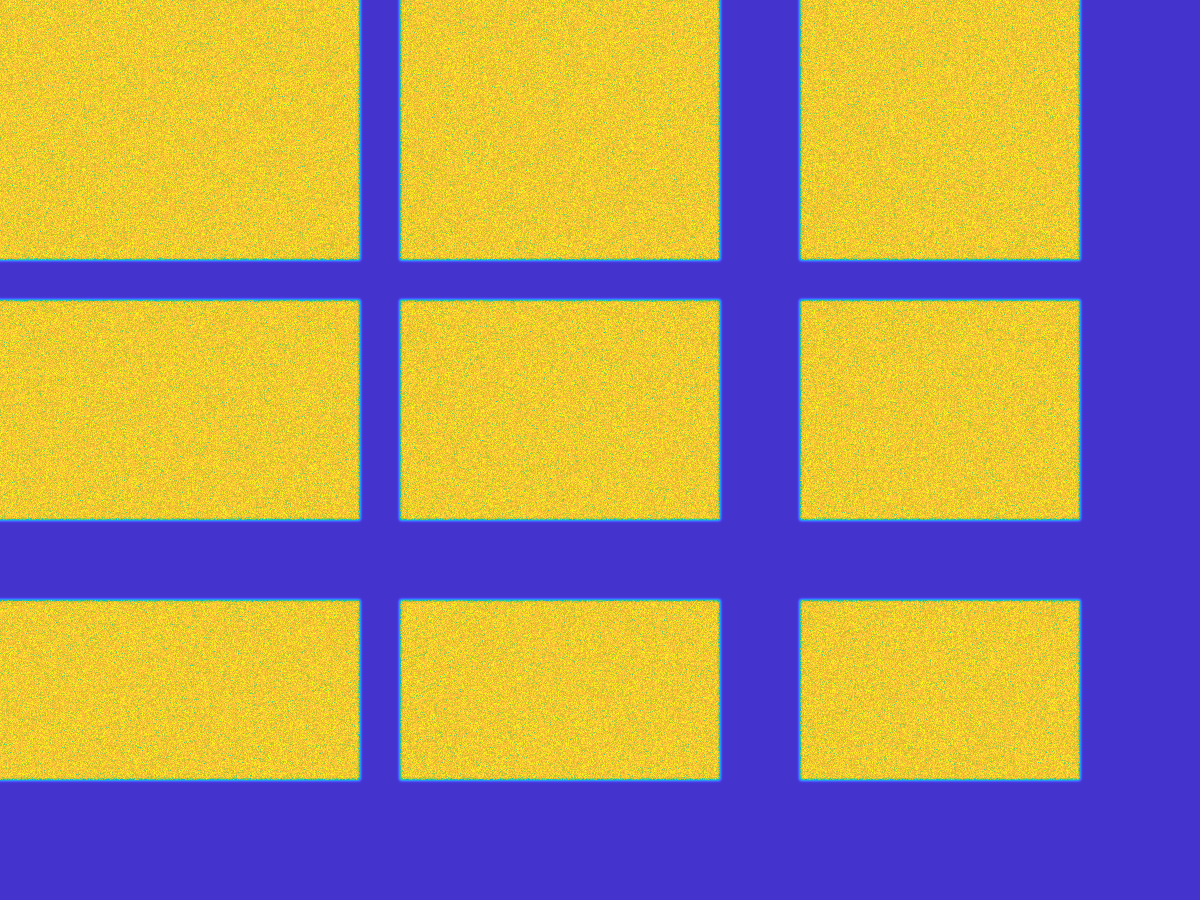} &
    \includegraphics[width=.31\linewidth]{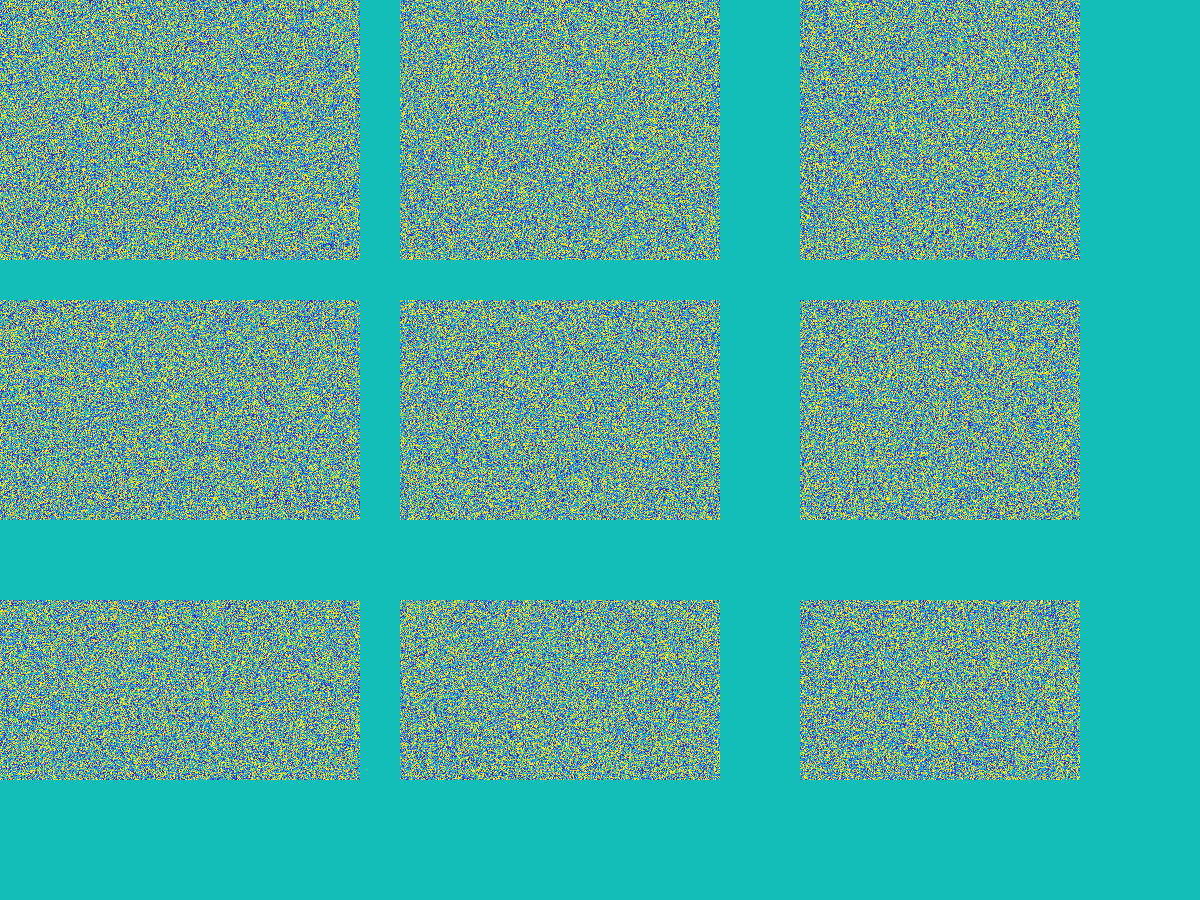} &
    \includegraphics[width=.31\linewidth]{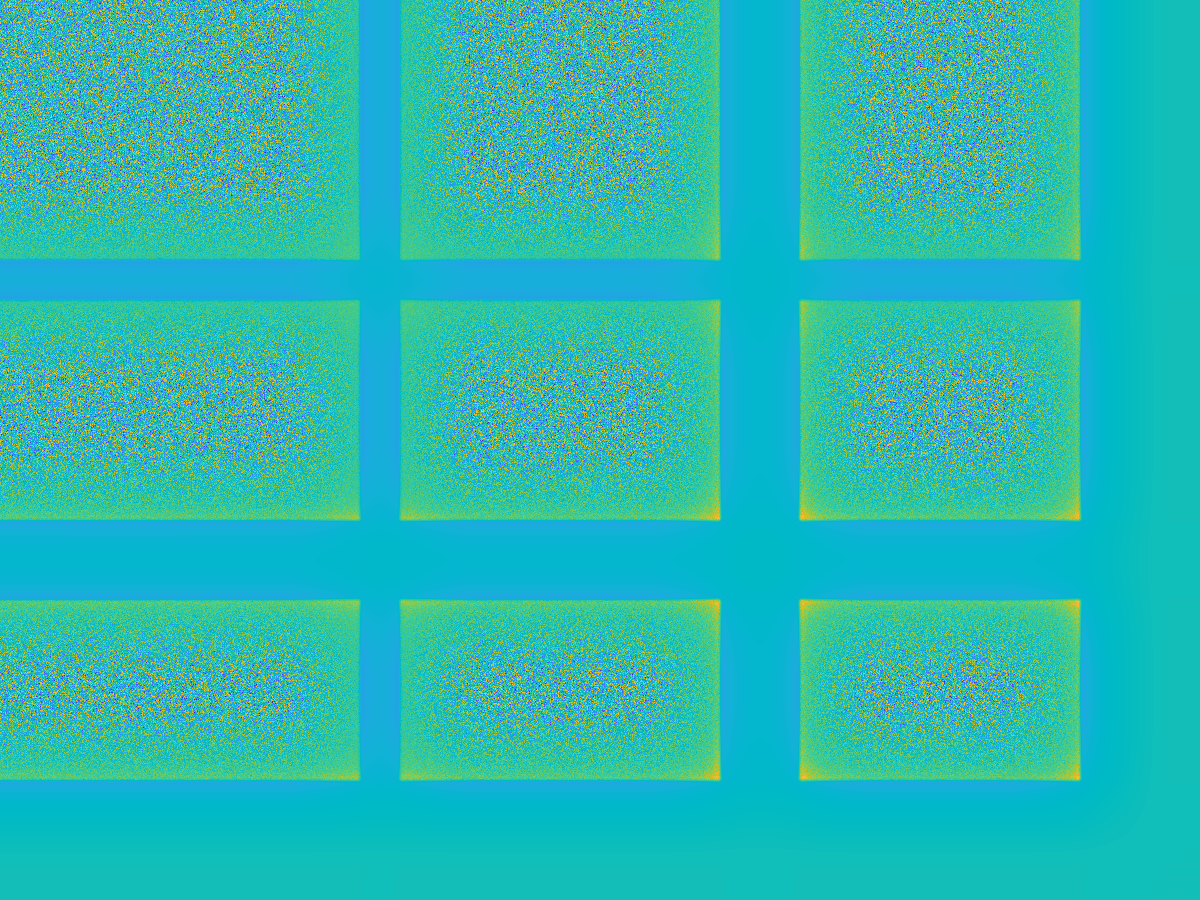} \\
    \textfg{(a) input test pattern} &
    \textfg{(b) expected detail} &
    \textfg{(c) detail with \ffgf{}}
  \end{tabular}
  \caption[Contrast halo test pattern and its worst filter]{
    Contrast halo pattern~(a); expected detail layer~(b), which
    contains only the noise of the test-pattern; result obtained with
    \ffgf{}~(c), which gives the strongest contrast halo score. All images
    are displayed with false colors. Dynamic range is $[0,1]$ for~(a) and
    $[-0.1,+0.1]$ for~(b) and~(c).}
  \label{fig:patterns:worstconthalo}
\vspace*{\floatsep}
  \begin{tabular}{cc}
    \includegraphics[width=.664\linewidth]{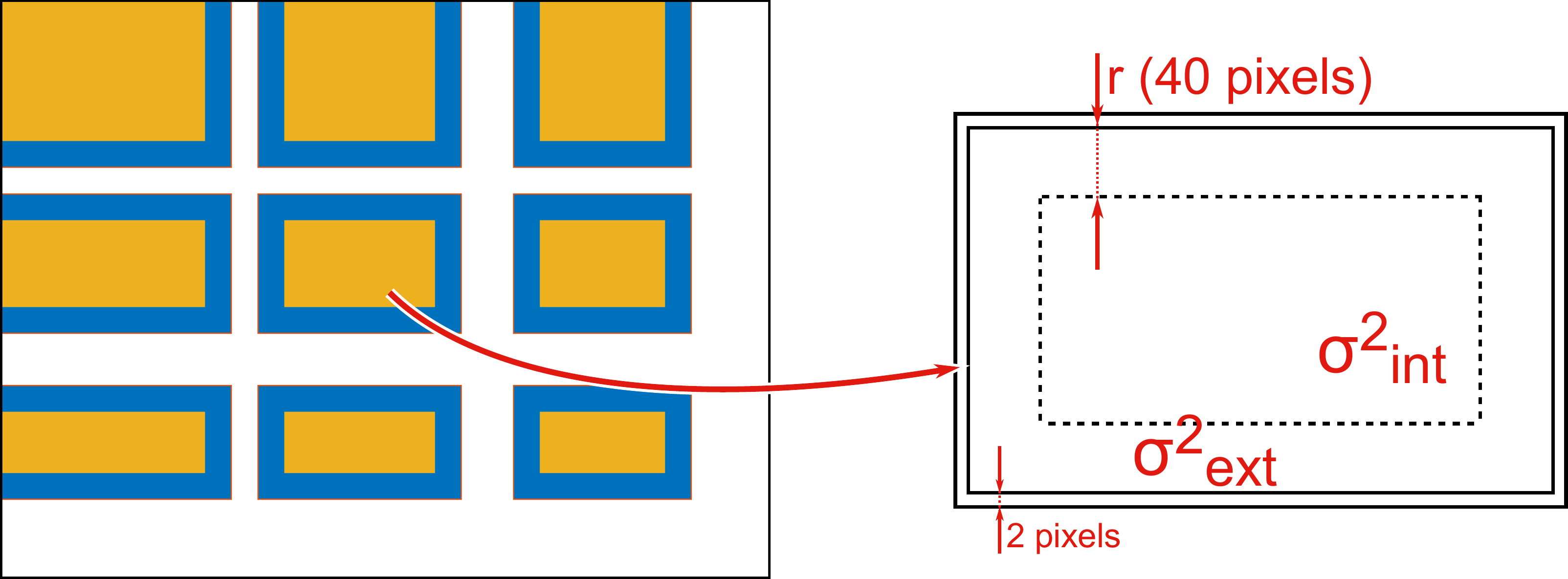}
  \end{tabular}
  \caption[Masks used in the contrast halo measure]{
    Masks used in the contrast halo measure.
      The variance $\sigma_{\text{ext}}^2$ is computed in the blue areas and
      the variance $\sigma_{\text{int}}^2$ in the yellow ones.
      The red lines show the excluded pixels on the borders of the rectangles.
      The white area is not used in the measure.}
  \label{fig:scheme:conthalo}
\end{figure} 

\paragraph{Test-pattern for the contrast halo}
This test-pattern  consists in a texture (noise) surrounded by contrasted edges
with different widths. We display it in Figure~\ref{fig:patterns} and with
false colors in Figure~\ref{fig:patterns:worstconthalo}~(a).

\paragraph{Measure}
Using the test-pattern described above, the contrast halo is measured by
comparing the variance of the detail layer in the interior of the bright
rectangles with the variance on the border of these rectangles, as shown in
Figure~\ref{fig:scheme:conthalo}. Because of the luminance halo, this ratio can
sometimes be inferior to 1, \ie{}, the variance in the exterior side of the
bright rectangles becomes higher than in the interior. We thus simply measure
the maximum of 1 and  of their ratio.
Formally, this gives

\begin{equation} \label{equ:measure:conthalo}
  \mathcal{C}(\im_0,\im_1) = \max\left\{1,
    {\sigma_{\text{int}}^2}/{\sigma_{\text{ext}}^2}\right\} - 1,
\end{equation}%
where the subtraction of $1$ only aims at giving the same minimum to
$\mathcal{C}$ as to the other measures, which will be useful in the final
comparison. The two measures of variances are obtained thanks to masks,
displayed in Figure~\ref{fig:scheme:conthalo}. The value of
$\sigma_{\text{ext}}^2$ is measured in the blue regions and
$\sigma_{\text{int}}^2$ in the yellow ones.

\paragraph{Filters with worst contrast halo}
The guided filter is the only filter among the tested ones that have a contrast
halo artifact.
The Figure~\ref{fig:patterns:worstconthalo}~(c) shows the filtering
result on this dedicated test-pattern: the texture is hardly removed near the
dark barriers. We provide as reference a synthetic detail layer we expect
  to be extracted by the filters. It contains only the noise of the
  test-pattern~(a).

\begin{figure*} 
  \centering
  \setlength{\tabcolsep}{.003875\linewidth}
  \begin{tabular}{@{}ccccc@{}}
    \includegraphics[width=.153\linewidth]{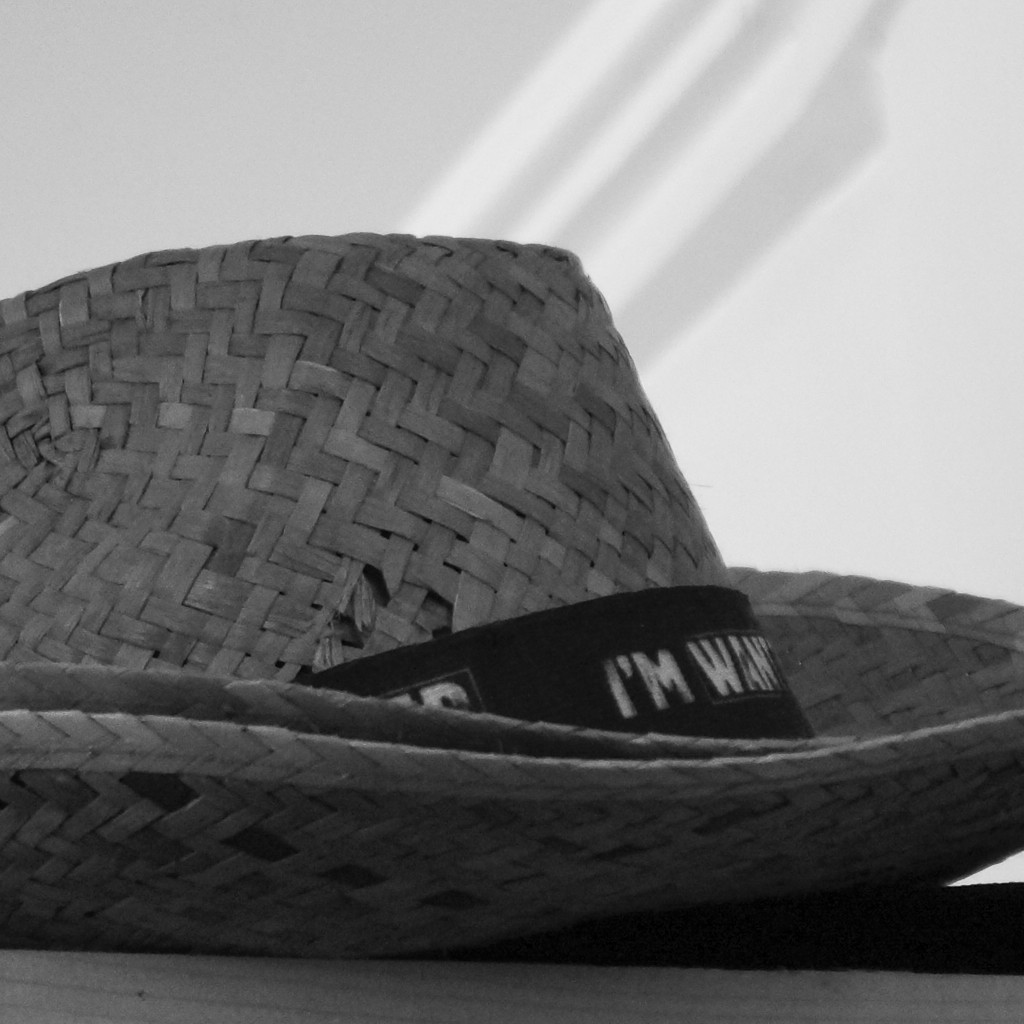} &
    \includegraphics[width=.153\linewidth]{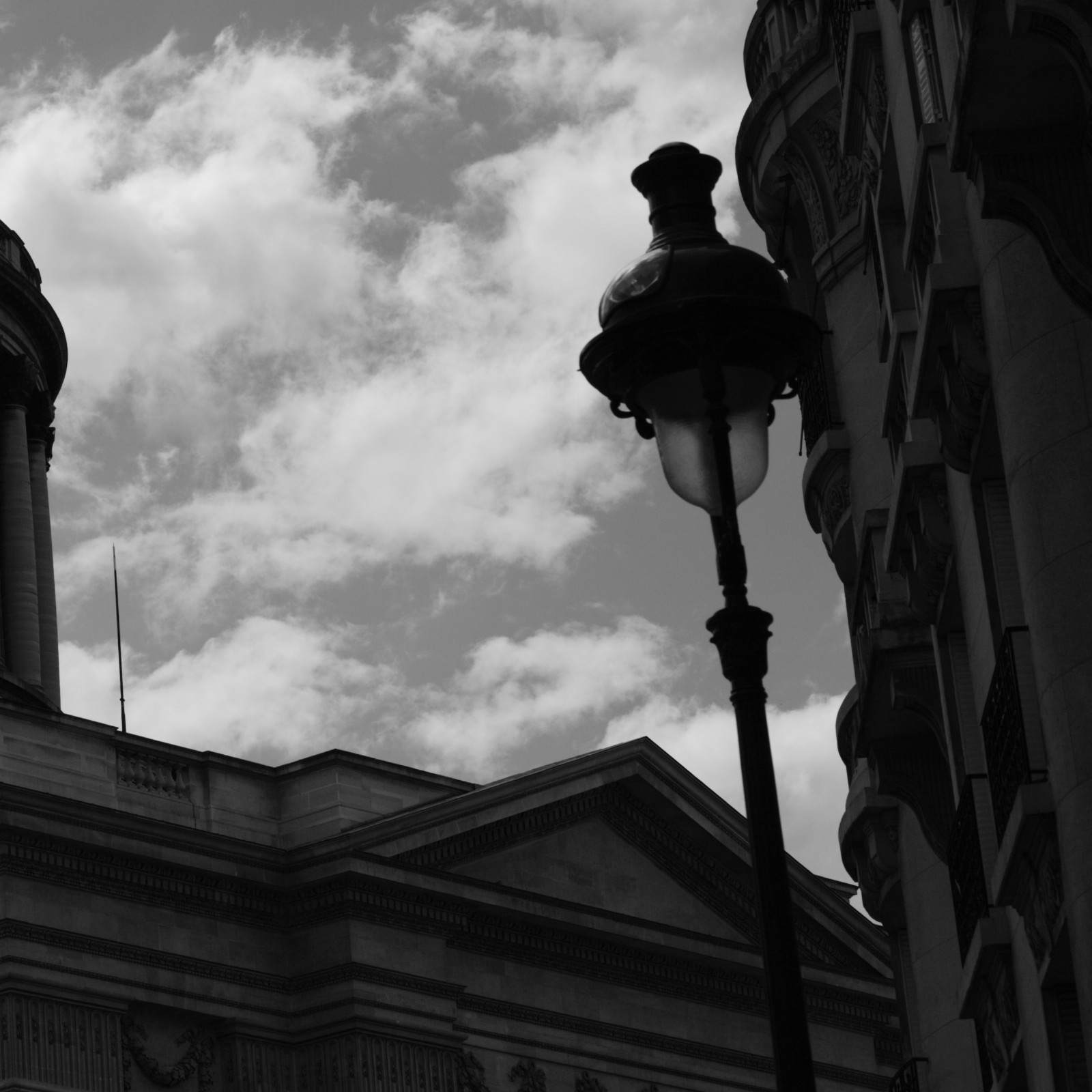} &
    \includegraphics[width=.153\linewidth]{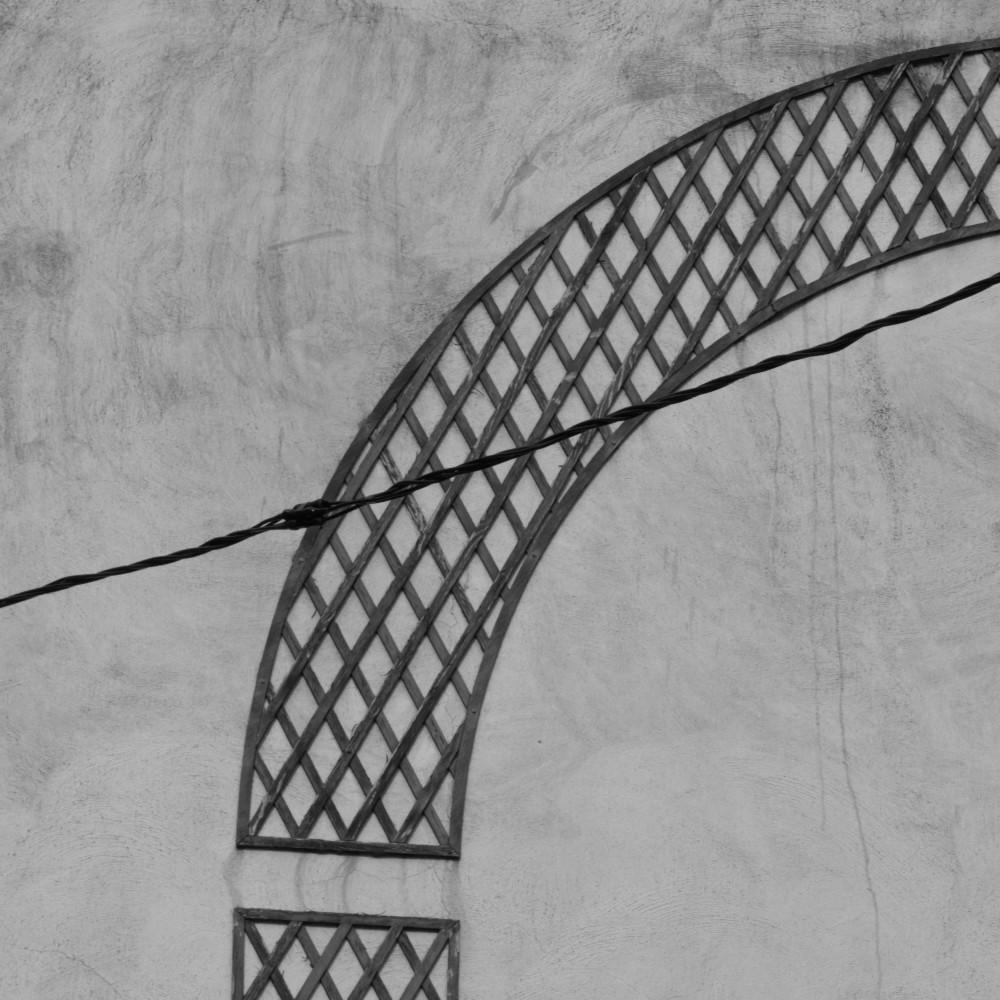} &
    \includegraphics[width=.204\linewidth]{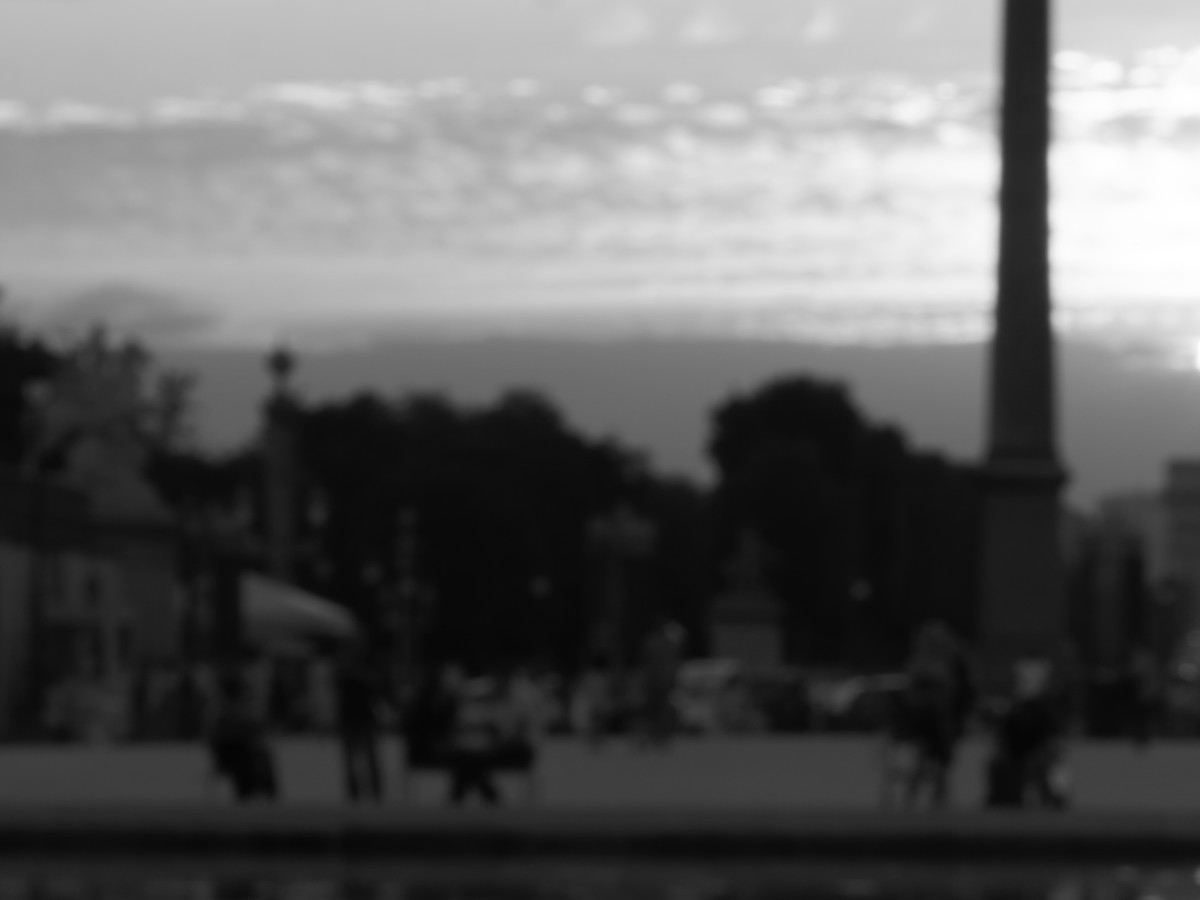} &
    \includegraphics[width=.306\linewidth]{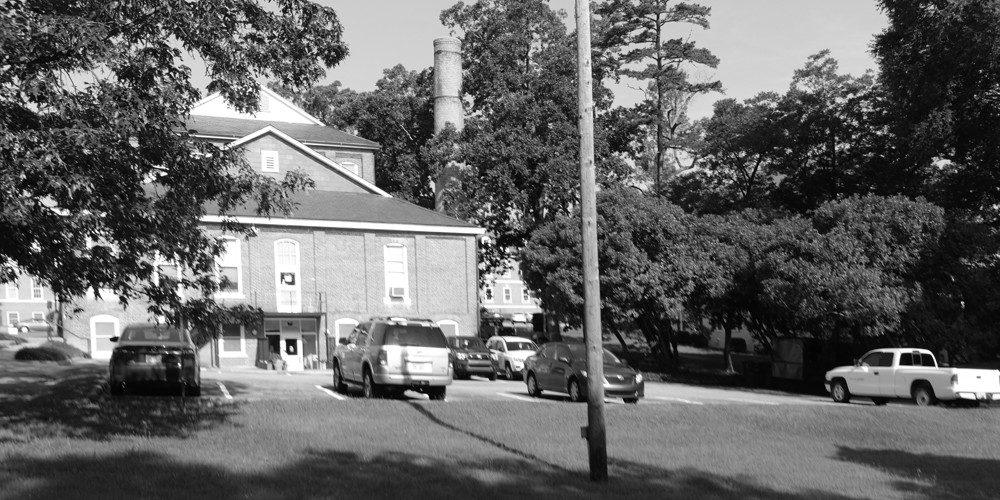}
  \end{tabular}
  \caption[Images used to set the parameters of all methods]{
    Images used to set the parameters of all methods.}
  \label{fig:testimages}
\end{figure*} 

\section{Setting the parameters of each filter}
\label{sec:params}

\begin{table*} 
  \centering
  \caption[List of the filters in competition and of their parameters]{%
    Tested filters and their parameters}
  \label{table:params}
  \begin{tabular}{|ll|l|l|}
    \hline
    {Abbreviation} &
    {Algorithm} &
    {Fixed} &
    {Set} \\ \hline
    \ffgf{} &
    Guided filter
    \cite{he2010guided, he2013guided} &
    $r = 40$ &
    $\epsilon = .075^2$ \\
    \fffbf{} &
    Fast bilateral filter (bilateral grid)
    \cite{paris2009fast, chen2007real}. Number of intensity samples: $S=64$ &
    $\sigma_s = 40$ &
    $\sigma_r = .1178$ \\
    \ffdt{} &
    Domain transform
    \cite{gastal2011domain}. Using the recursive filter and $3$ iterations &
    $\sigma_s = 40$ &
    $\sigma_r = 0.1166$ \\
    \ffwls{} &
    Weighted Least Squares
    \cite{farbman2008edge}. Guide image $\ell$ set to default: log-luminance
    of input $\im$ &
    $\alpha = 1.2$ &
    $\lambda = 0.5$ \\
    \fffllf{} &
    Fast local Laplacian filter 
    \cite{aubry2014fast}. Number of samples for range subsampling: $S=64$ &
    $l_{\text{\tiny max}}=6$ &
    $\sigma_r = 0.103$ \\
    \fftv{} &
    Total variation using $L^1$ norm
    \cite{guen2014cartoon} &
    - &
    $\lambda = .205$ \\
    \fflois{} &
    $L^0$ image smoothing
    \cite{xu2011image} &
    $\kappa = 2$ &
    $\lambda = .002$ \\
    \hline
  \end{tabular}
\end{table*} 

In this section we discuss how to set fair parameters for each filter.
When we shall compare in Section~\ref{sec:measures} the strengths of
their respective artifacts, it is of prime importance to ensure that they are
compared in  a condition where they deliver similar contrast enhancement.

Using the default parameters given by the authors wouldn't be right. Notably, a
cautious filter might cause less visible artifacts, but the detail enhancement might be insufficient by then.
Our problem is thus to obtain similar amounts of detail.

We therefore propose to equalize the $L^2$-norm of the detail layer.

The methodology we propose can be characterized as semi-automatic. For each
filter, we set all parameters using two basic rules, and the last and more
important one is set automatically.
The rules are simply (1) to use default values suggested by the authors
when possible, while (2) ensuring a coherence between the different filters. For
example, if available, parameters that control the spatial scale of the detail extraction
are set to the same value. The last parameter set automatically is the one that
controls the amount of detail. This
parameter is set so that the average $L^2$-norm of the detail layer on a
small set of images is the same for each filter.
The five images we used are displayed in Figure~\ref{fig:testimages}.
They were chosen so as to present real-life cases with a standard balance of texture and
  edges.

To avoid penalizing smaller images, we preferred to equalize the \psnr{} (Peak
Signal to Noise Ratio) of the images, because it is independent of their size,
\begin{equation} \label{equ:comp:measure:psrn}
  \psnr( \im_0 - \im_1 ) = - 10 \times \log_{10}\left( \frac{1}{\sqrt{N}}
    \| \im_0 - \im_1 \|_{2} \right),
\end{equation}
where $N$ is the number of pixels in the image.
The \psnr{} was measured on each image filtered with the current parameter, then
averaged. For the  cross-calibration of all filters we fixed the target at $\psnr\!=\!16.23$~dB, which corresponds to a decent amount of detail for all filters.

We list in Table~\ref{table:params} the fixed parameters and give in the last
column the parameters obtained with our procedure\footnote{%
  The detail layers obtained with these parameters for the images in
  Figure~\ref{fig:testimages} are presented in the supplementary material.}.

\section{Measuring the  artifacts}
\label{sec:measures}

\begin{figure*} 
  \centering
  \setlength{\tabcolsep}{.010\linewidth}
  \begin{tabular}{@{}cccc@{}}
    \includegraphics[width=.235\linewidth]{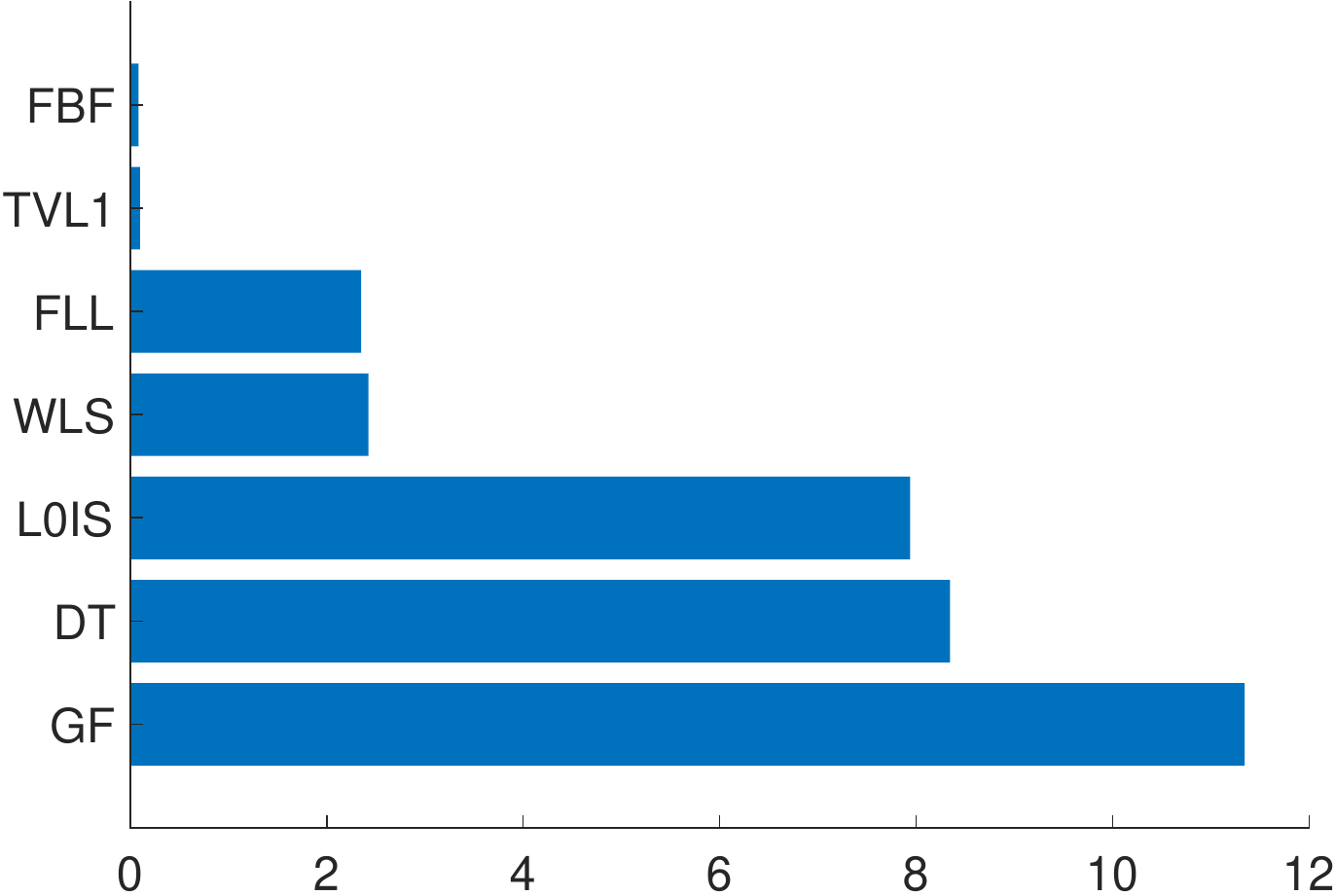}&
    \includegraphics[width=.235\linewidth]{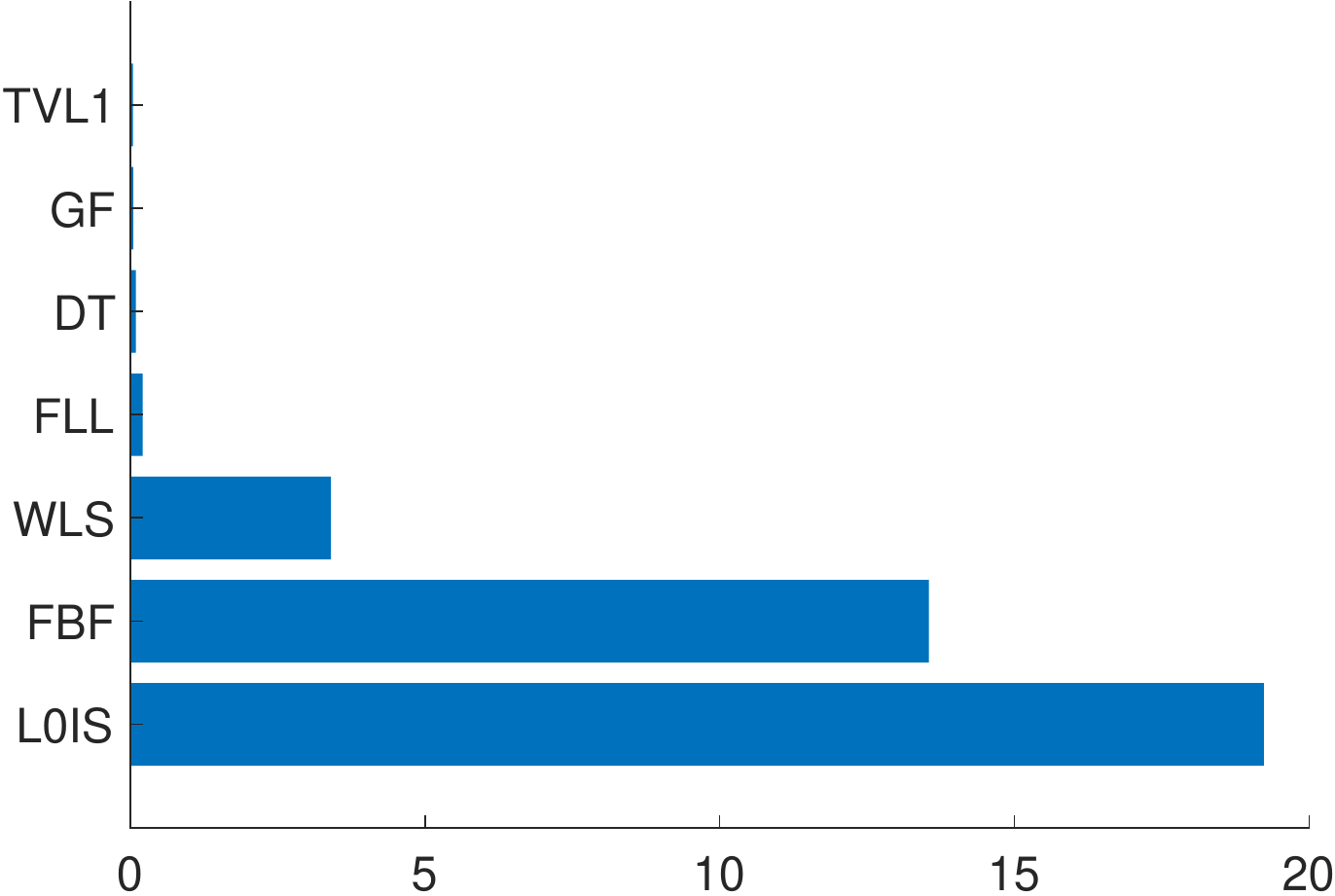}&
    \raisebox{-.075\height}{\includegraphics[width=.235\linewidth]{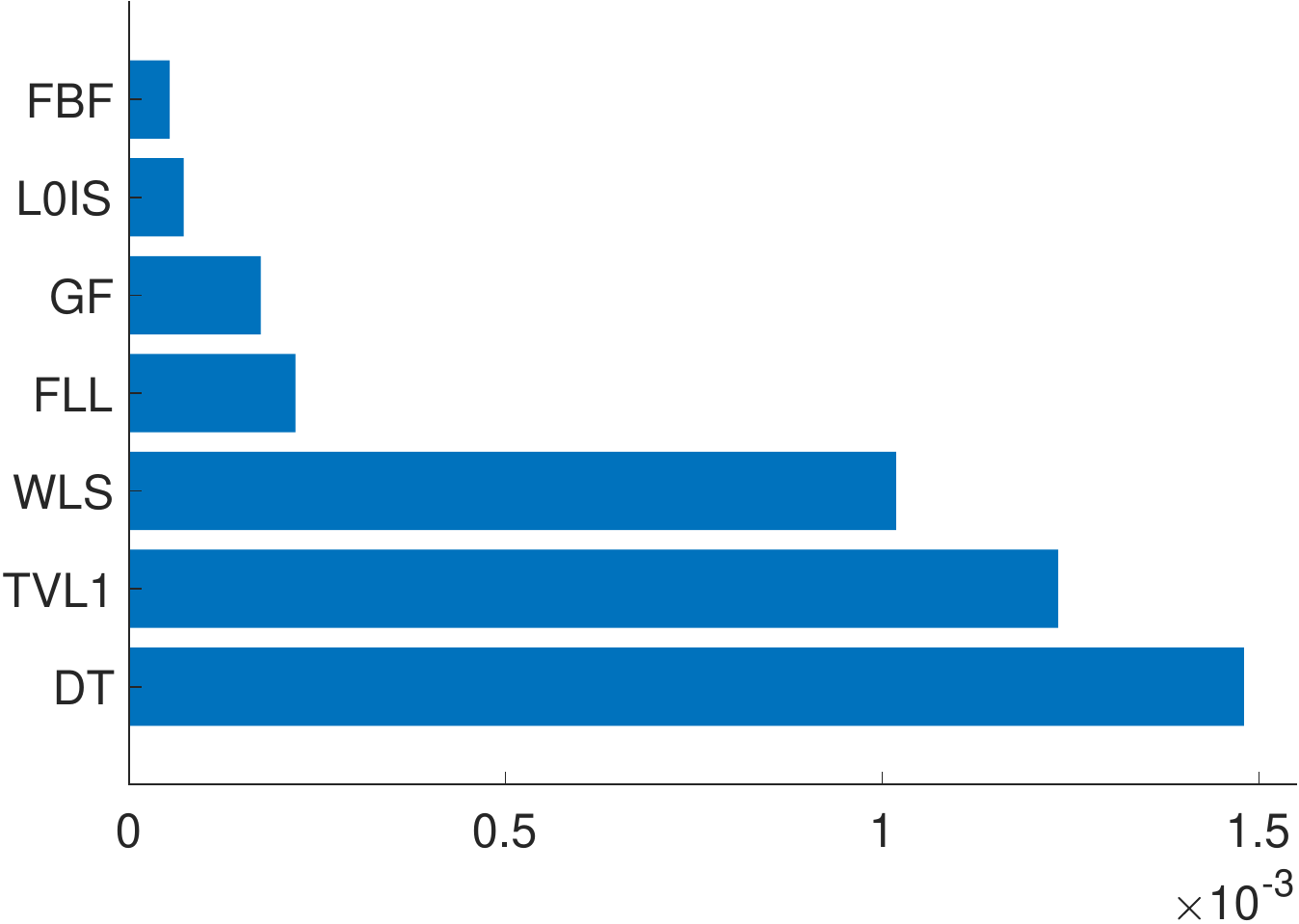}}&
    \includegraphics[width=.235\linewidth]{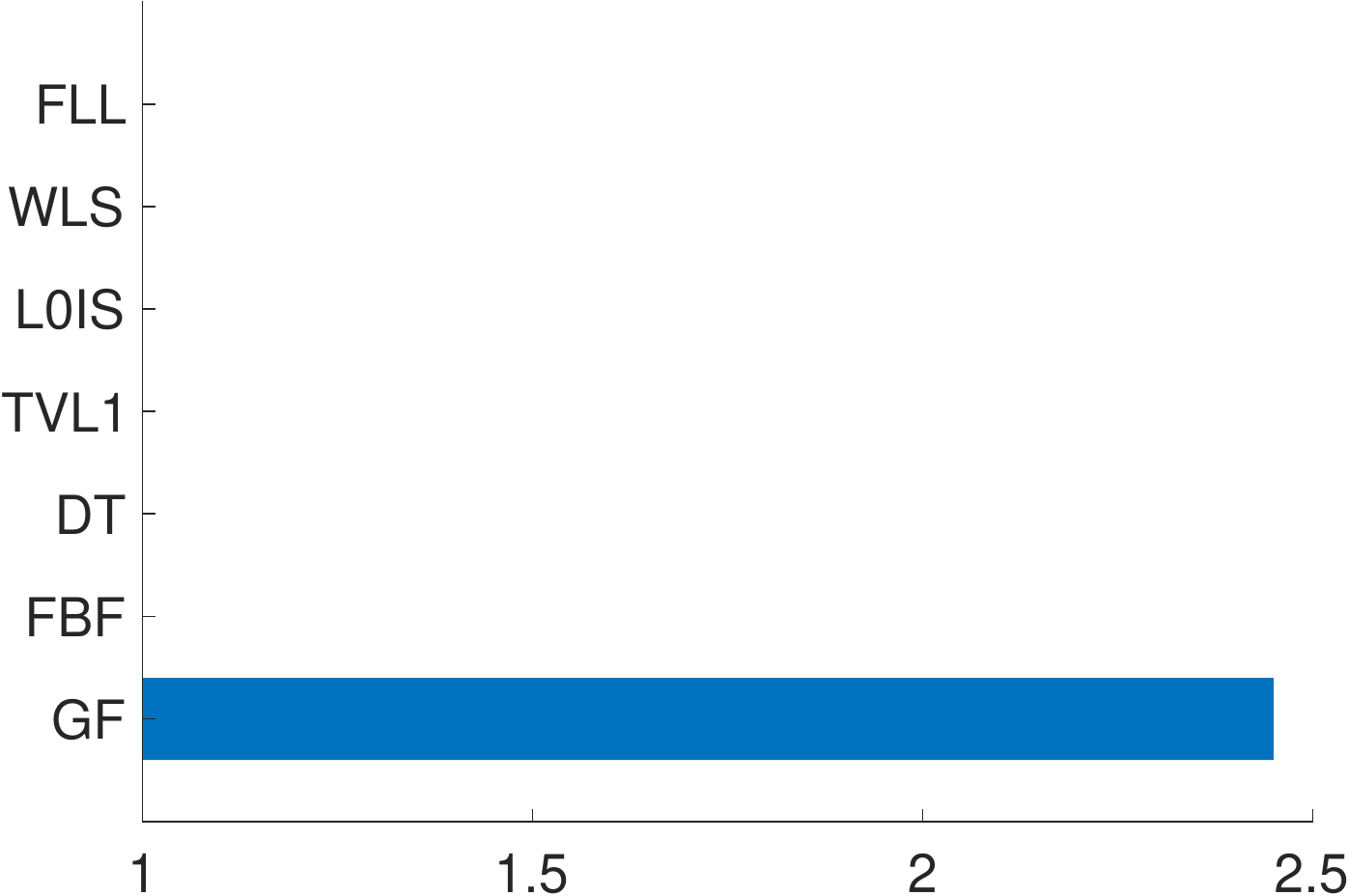}\\
    \textfg{(a) luminance halo ($\mathcal{H}$)} &
    \textfg{(b) staircasing ($\mathcal{S}$)} &
    \textfg{(c) compartmentalization ($\mathcal{P}$)} & 
    \textfg{(d) contrast halo ($\mathcal{C}$)}
  \end{tabular}
  \caption{Objective measures of the artifacts. For each plot the filters
      are ranked according to their score. The lower the better.}
  \label{fig:measures}
\end{figure*} 

In this section, we measure the strength of each of the four artifacts presented
above. A table will display the results for all the filters listed above.
Our method is simple: for each filter, using the parameters given in
Table~\ref{table:params}, we measured the tested artifact using equations
given in Section~\ref{sec:testpatterns}.
We give in Figure~\ref{fig:measures} four bar plots presenting the values
$\mathcal{H}$, $\mathcal{S}$, $\mathcal{P}$ and $\mathcal{C}$ for each filter,
sorted in descending order\footnotemark{}. The smaller the value, the better
the filter; this way the filters are directly ranked.

\footnotetext{\label{footnote:suppmat:detailpatterns}%
  The detail layers obtained with the different filters for each
  pattern are observable in the supplementary materials.}

In the first Section~\ref{subsec:measures:artifacts}, we comment and explain the results.
  In a second Section~\ref{subsec:measures:final}, we make a summary of the five tables.
We propose a method to merge the independent scores into a single value that
summarizes the ability of the filter to perform a clean base+detail
decomposition and deliver the final ranking. This  ranking is confirmed by a
rejection table summarizing the experts' evaluation of artifacts.
The last Section~\ref{subsec:measures:conclusion} summarize and concludes on our
results.

\subsection{Artifact-wise measures and ranking}
\label{subsec:measures:artifacts}

\paragraph{Luminance halo}
The worst filters in this case are \fflois{} and   \ffgf{}\footnotemark{}.

\footnotetext{See Footnote~\ref{footnote:suppmat:detailpatterns}}

\paragraph{The  staircase effect}
Once again
\fflois{} is the worst filter\footnotemark{}: indeed, minimizing the $L^0$~norm of the
gradients tends to create constant parts in the image, which creates a staircase effect. Unsurprisingly, the (fast) bilateral filter comes next
after \fflois{}. This artifact of the bilateral filter has long been
  known \cite{buades2006staircasing}.
The weighted least squares filter has, to a certain extent,
this artifact too, with the particularity that it is way more marked in the dark
side of the halo. This is due to the use of a logarithm in the gradient
penalization: dark parts of the images are allowed to move more than the bright
ones.

\footnotetext{See Footnote~\ref{footnote:suppmat:detailpatterns}.}

\paragraph{The  compartmentalization artifact}
The expected detail is a constant image: indeed, the test-pattern does not
contain texture, but only very contrasted edges.
On the contrary, filters with compartmentalization tend to darken the background
stripes separating the rectangles and to ``light up'' some of
the squares or rectangles of the test-pattern in function of their area.
The worst result for this measure is obtained with the domain transform (\ffdt{}), that tends
to smooth out the small objects whatever their contrast\footnotemark{}.
Note that contrarily to \ffwls{} that lights up only some shapes in function of
their areas, \ffdt{} affects every shape. So for this filter the
compartmentalization is linked to the luminance halo.
As expected, \ffwls{} has a very high score too.
This  is explainable because even the edges see their
gradients slightly penalized. Thus it becomes sometimes worthy, in terms of energy
minimization, to reduce those edges if the area inside is small enough, because
the data term, having few pixels, will not compensate the gain. The second
worst filter is \fftv{}. Indeed, this filter is prone to removing the edges of
objects and ``closing'' regions with small area.
Next, \fflois{}  also has a bad score, for the same reason as presented in the
luminance halo paragraph. In its case the compartmentalization is not really
annoying because it seems to affect the shapes whatever their area. Other
non-zero results are mainly due to the luminance halo, to which our test-pattern cannot
be completely insensitive. Note also that the ``contour highlighting'' visible
in \fffbf{}, \ffbfr{} and \fflois{} is due to the staircase effect. This, however,
does not influence the value of our measurement.

\footnotetext{See Footnote~\ref{footnote:suppmat:detailpatterns}.}

\paragraph{The contrast halo results}
As explained in
Section~\ref{sec:testpatterns}, with this test-pattern we aim at measuring
if the detail is affected in the  vicinity of  an edge. Put another way, we measure
if the smoothing is the same in the vicinity of edges as at a certain distance
from it.
The filter that obtains the highest and therefore worst score is the
guided filter\footnotemark{}.
Its detail layer is displayed in
  Figure~\ref{fig:patterns:worstconthalo}, along with the guided filter's
  result.

\footnotetext{See Footnote~\ref{footnote:suppmat:detailpatterns}.}

\subsection{Final score and ranking}
\label{subsec:measures:final}

Merging the different measures may look problematic for three main reasons:
\begin{enumerate}
  \item the perceived nuisance of an artifact is non-linear;
  \item the artifacts are not equally disturbing;
  \item our measures have ranges that are not comparable.
\end{enumerate}

When global quality assessment scores are available,
one can overcome these difficulties by performing a polynomial regression.
This method has been used for example in the context of tone-mapped images
quality assessment \cite{cadik2006image, cadik2008evaluation} to merge different
quality measures and try to approximate the subjective response of a large
number of non-expert subjects\footnotemark{}.

\footnotetext{See the supplementary material.}

We, however, worked with five experts  at \dxo{}, as the evaluation given by
photography experts is much more reliable. It actually also relies  on the
feedback of many amateur photographers.
Rather than attributing global notes to the filters,
we intend to detect objectionable artifacts in the filters.
This rating method is similar to
\cite{trentacoste2012unsharp, aydin2008dynamic}, that identified objectionable
thresholds using perceptual studies.

These experts were asked for a clear cut categorization of each triplet (image,
filter, artifact) in two classes: \emph{non-objectionable}, and
\emph{objectionable}.
The experiment was carried-out on contrast enhanced images with a high factor on
the detail. The comparison was performed by flipping between the input images
provided for reference and the enhanced ones.
The set of images was partly composed of the examples displayed in
Figure~\ref{fig:testimages}, completed with 10 other challenging
images from \dxo{} database.
They were asked in the end to make a decision for the couple (filter, artifact)
that summarizes their observations,
so that we obtained a rejection table for every expert.
We then merged the tables.
To secure an objectionable  decision, we marked the couple (filter, artifact) as
rejected only if at least four of the five judges marked them as objectionable.
We provide in Table~\ref{table:measures:unacceptables} the result of their
judgment.

\begin{figure} 
  \centering
  \begin{tabular}{@{}c@{}}
    \includegraphics[width=\linewidth]{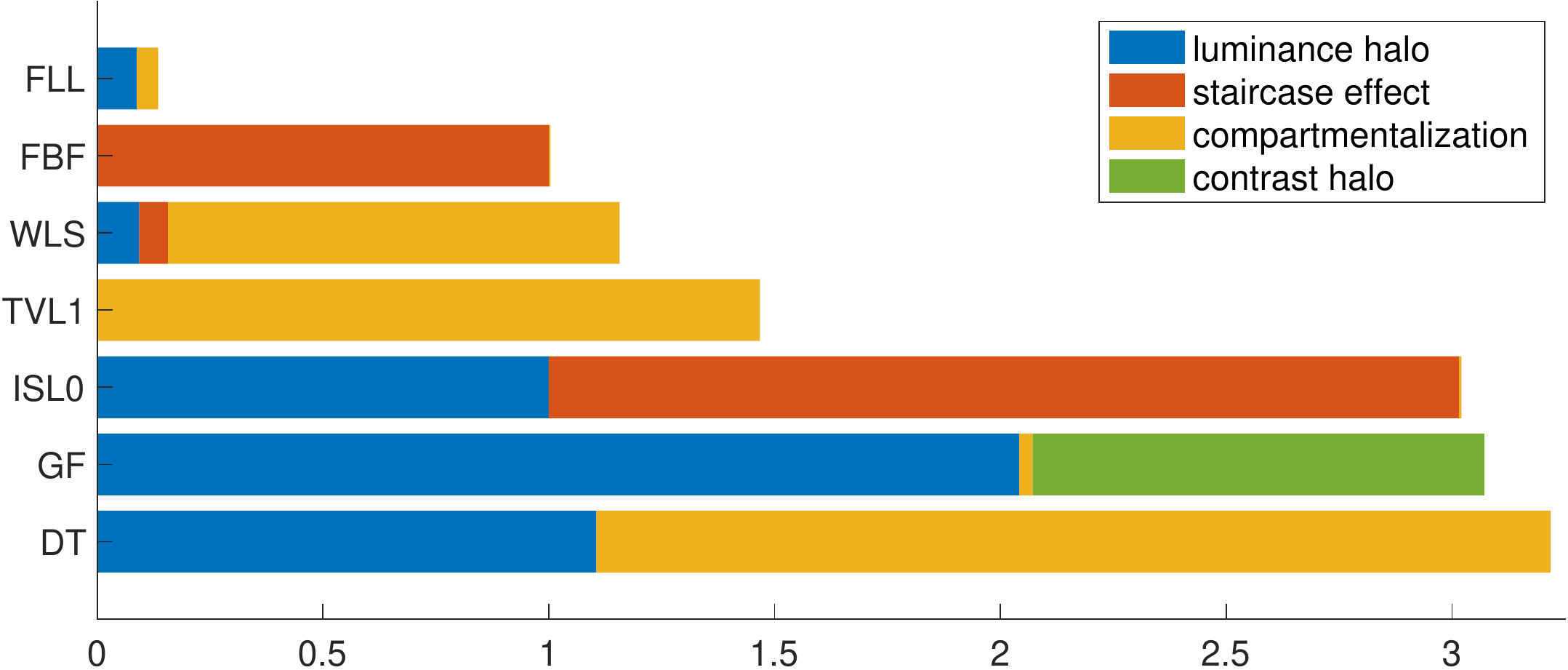} \\
  \end{tabular}
  \caption[Final comparison of the measured values]{%
    Comparison and ranking of the fused scores $\mathcal{A}$ obtained with
    Equation~\eqref{equ:measures:fusionVtwo}. The lower the better.}
  \label{fig:measures:finalrankingVtwo}
\end{figure}

For each tested filter, Table~\ref{table:measures:unacceptables} fixes its
objectionable artifacts.  We shall only use this expert rating  to fix a
threshold for the artifact's score. The value we seek lies between the largest
non-objectionable score and the smallest objectionable one.
We chose to use the smallest objectionable score as threshold for each artifact. This score is therefore associated with one of our seven filters.
Hence, using our measures in Figure~\ref{fig:measures} and 
Table~\ref{table:measures:unacceptables}, we have that
\begin{itemize}
  \item \fflois{} defines the score threshold for the luminance halo,
  \item \fffbf{} defines the  score threshold for the staircasing,
  \item \ffwls{} defines the score threshold for the compartmentalization,
  \item \ffgf{} defines the  score threshold for the contrast halo.
\end{itemize}
The scores of these filters are then used to normalize the measures.
This provides a workable solution to the two last difficulties listed above. Concerning
the first difficulty, we chose to square the artifact measures. This makes the objectionable scores more
discriminant and the acceptable ones less impacting, and translates in the simple fusion equation:
\begin{equation} \label{equ:measures:fusionVtwo}
  \mathcal{A}(f) = \frac{\mathcal{H}(f)^2}{\mathcal{H}(\fflois{})^2}
                 + \frac{\mathcal{S}(f)^2}{\mathcal{S}(\fffbf{})^2}
                 + \frac{\mathcal{P}(f)^2}{\mathcal{P}(\ffwls{})^2}
                 + \frac{\mathcal{C}(f)^2}{\mathcal{C}(\ffgf{})^2},
\end{equation}
where $f$ is a filter and $\mathcal{A}(f)$ its final score taking into account
the four artifacts.
The fused measures are compared (and sorted) in
Figure~\ref{fig:measures:finalrankingVtwo}.

\begin{table*}
  \centering
  \caption[Unacceptable artifacts]{%
    Unacceptable artifacts (in red) in the tested filters.}
  \label{table:measures:unacceptables}
  \begin{tabular}{|r|%
      >{\centering\arraybackslash}m{.05\linewidth}%
      >{\centering\arraybackslash}m{.05\linewidth}%
      >{\centering\arraybackslash}m{.05\linewidth}%
      >{\centering\arraybackslash}m{.05\linewidth}%
      >{\centering\arraybackslash}m{.05\linewidth}%
      >{\centering\arraybackslash}m{.05\linewidth}%
      >{\centering\arraybackslash}m{.05\linewidth}|}
    \hline &
    \textfg{\ffdt{}} &
    \textfg{\fflois{}} &
    \textfg{\fffbf{}} &
    \textfg{\fffllf{}} &
    \textfg{\ffgf{}} &
    \textfg{\fftv{}} &
    \textfg{\ffwls{}} \\ \hline
    \raisebox{-.667\height}{{staircase effect}} &
    \cellcolor{YellowGreen!80} &
    \cellcolor{Red!80} &
    \cellcolor{Red!80} &
    \cellcolor{YellowGreen!80} &
    \cellcolor{YellowGreen!80} &
    \cellcolor{YellowGreen!80} &
    \cellcolor{YellowGreen!80} \\[2ex]
    \raisebox{-.667\height}{{compartmentalization}} &
    \cellcolor{Red!80} &
    \cellcolor{YellowGreen!80} &
    \cellcolor{YellowGreen!80} &
    \cellcolor{YellowGreen!80} &
    \cellcolor{YellowGreen!80} &
    \cellcolor{Red!80} &
    \cellcolor{Red!80} \\[2ex]
    \raisebox{-.667\height}{{contrast halo}} &
    \cellcolor{YellowGreen!80} &
    \cellcolor{YellowGreen!80} &
    \cellcolor{YellowGreen!80} &
    \cellcolor{YellowGreen!80} &
    \cellcolor{Red!80} &
    \cellcolor{YellowGreen!80} &
    \cellcolor{YellowGreen!80} \\[2ex]
    \raisebox{-.667\height}{{luminance halo}} &
    \cellcolor{Red!80} &
    \cellcolor{Red!80} &
    \cellcolor{YellowGreen!80} &
    \cellcolor{YellowGreen!80} &
    \cellcolor{Red!80} &
    \cellcolor{YellowGreen!80} &
    \cellcolor{YellowGreen!80} \\[2ex] \hline
  \end{tabular}
\end{table*}

\subsection{Summary and conclusion on the comparative experiments} 
\label{subsec:measures:conclusion}

We find in the fused scores in Figure~\ref{fig:measures:finalrankingVtwo} that
the fast local Laplacian filter (\fffllf{}) is the best and the domain transform
(\ffdt{}), image smoothing with $L^0$ gradient minimization (\fflois{}) and the
guided filter (\ffgf{}) are the worst of the tested methods.

This is in excellent agreement with the experts' rejection decisions.
In Table~\ref{table:measures:unacceptables}, each test-pattern disqualifies at
least one method:
\begin{itemize}
  \item the staircase effect invalidated \fflois{} and \fffbf{};
  \item the compartmentalization invalidated \ffdt{}, \fftv{}, \ffwls{};
  \item the contrast halo invalidated \ffgf{};
  \item the luminance halo invalidated \ffdt{}, \ffgf{}, and \fflois{}.
\end{itemize}
Only \ffllf{} succeeded passing the five artifact tests.
Thus, this classification confirms the podium obtained in
Figure~\ref{fig:measures:finalrankingVtwo}, where the first place is attributed
to the same filter.
Furthermore, the 2\textsuperscript{nd}, 3\textsuperscript{rd} and
4\textsuperscript{th} filters are the ones with only one objectionable artifact.
All remaining filters suffer from two intolerable artifacts and get the last
positions in the ranking.

\section{Conclusion}

The emergence of more and more effective TMOs, applied to the richer content of
an HDR image has  introduced new degrees of freedom in image rendering.
One of the sources of this incredible flexibility is the recent invention of
powerful base and detail
decomposition filters, that lie at the heart of local contrast manipulation
methods.  The multiplication of methods requires a quality measurement, independent of questionable aesthetic criteria.

We proposed an objective measure of the presence of artifacts in the base and
detail decomposition filters. This involves several important contributions: the
definition and measurement of four artifacts using
\textit{ad hoc} test-patterns, a methodology to set the parameters of the filters
and the proposition of an unprecedented way to  fuse the independent measures
in a single quality mark.
Our procedure evaluates the filters according to criteria that faithfully
reflect the photographers' requirements and are easily applicable to any base+detail filter.

Beyond the ranking of existing methods, we believe that our protocol can serve
the design of new filters and the tuning of their parameters. The quantitative
evaluation of success is a simple tool that can validate of invalidate any new
proposed method.

Although we obtained for each artifact an expert acceptability threshold, we believe
that our evaluation could become still more precise with a perceptual evaluation of a greater
breadth.

An interesting direction for future works would be to extend our method to all
contrast enhancement filters, not only those that explicitly produce a base and
a detail layer. The difficulty remaining is to define an objective
cross-calibration applicable to all such filters, so as to compare them all for
a prefixed amount of enhancement.

\section*{Acknowledgments}
Work partly financed by Office of Naval research  grant N00014-17-1-2552,   DGA
Astrid project \guillemotleft{} filmer la Terre \guillemotright{} n\degree{}
ANR-17-ASTR-0013-01.
Charles Hessel's Ph.D. scholarship was supported by a CIFRE scholarship of the
French Ministry for Higher Studies,  Research and Innovation.
He wishes to thank his DxO collaborators Gabriele Facciolo, Wolf Hauser, Quoc
Bao Do, Beno\^{i}t Chauville, Fr\'{e}d\'{e}ric Guichard  for many valuable
conversations and advice.

\bibliographystyle{IEEEtran}
\bibliography{IEEEabrv,bibliography/biblio.bib}

\begin{IEEEbiographynophoto}{Charles Hessel}
received the B.Sc. and M.Sc. degrees in eletronic, electrotechnic and automation
from University Toulouse III Paul Sabatier, Toulouse, France, in 2012
and 2014 respectively.
He obtained a CIFRE scholarship from the French Ministry of Higher Education,
Research and Innovation, and was a Ph.D. student at \dxo{},
Boulogne-Billancourt, France, and the Centre de Math\'{e}matiques et leurs
Application (CMLA), at the \'{E}cole Normale Sup\'{e}rieure Paris-Saclay,
Cachan, France, from which he received the Ph.D.  degree in applied mathematics
in 2018.
He is currently a postdoctoral researcher at the \'{E}cole Normale
Sup\'{e}rieure Paris-Saclay.
\end{IEEEbiographynophoto}
\begin{IEEEbiographynophoto}{Jean-Michel Morel}
received the PhD degree in applied mathematics from University Pierre et Marie
Curie, Paris, France in 1980. He started his career in 1979 as assistant
professor in Marseille Luminy, then moved in 1984 to University Paris-Dauphine
where he was promoted professor in 1992.  He is Professor of Applied Mathematics
at the \'{E}cole Normale Sup\'{e}rieure Paris-Saclay since 1997. His research is
focused on the mathematical analysis of image processing. He is a 2013 laureate
of the \textit{Grand Prix Inria-Acad\'{e}mie des Sciences}, a 2015 laureate of
the Longuet-Higgins prize, and of the 2015 CNRS
{\it m\'{e}daille de l’innovation}.
\end{IEEEbiographynophoto}
\end{document}